\documentclass[10pt]{article}
\usepackage[preprint]{tmlr}

\usepackage{amsmath,amsfonts,bm}

\def\eqref#1{equation~\ref{#1}}

\def\1{\bm{1}}

\DeclareMathAlphabet{\mathsfit}{\encodingdefault}{\sfdefault}{m}{sl}
\SetMathAlphabet{\mathsfit}{bold}{\encodingdefault}{\sfdefault}{bx}{n}

\usepackage{hyperref}
\usepackage{url}
\usepackage{booktabs}
\usepackage{graphicx}
\usepackage{amsmath,amssymb}
\usepackage{xcolor}
\usepackage{tikz}
\usepackage{algorithm}
\usepackage[noend]{algpseudocode}

\usetikzlibrary{arrows.meta,positioning}

\newcommand{\method}{SYNAPSE}

\title{SYNAPSE: Sharing Tool-Use Experience\\Across Organizations with Differential Privacy}

\newcommand{\asu}{Arizona State University}
\author{%
  \begin{tabular}[t]{@{}l@{}}
    \name Abhijit Chakraborty \\
    \addr \asu \\
    {\small\texttt{achakr40@asu.edu}}
  \end{tabular}\hfill%
  \begin{tabular}[t]{@{}r@{}}
    \name Vivek Gupta \\
    \addr \asu \\
    {\small\texttt{vgupt140@asu.edu}}
  \end{tabular}}

\def\month{MM}
\def\year{YYYY}
\def\openreview{\url{https://openreview.net/forum?id=XXXX}}

\begin{document}
\maketitle

\begin{abstract}
An agent with a large tool catalog first retrieves a few candidate tools for each request. Examples of how people used each tool help this step, but usage records are private and stay with the organizations that hold them. We present \method{}, a way for organizations to share tool-use experience without sharing records. They jointly train a request generator under differential privacy (DP) and release noisy per-tool usage counts, which set how many requests the generator writes for each tool. Generated requests become each tool's usage field in a shared index, scored beside public documentation. For fixed scoring settings, each release of the mechanism is record-level DP if its noise stays secret, at least three of the five clients are honest and secure aggregation holds; choosing the settings on a second release from the same pool doubles the budget. Our benchmark simulates the mechanism with public, reproducible noise, so it measures utility and its outputs carry no privacy guarantee. We test \method{} with five hash-partitioned clients drawn from one benchmark pool, on two catalogs of real tools whose requests were written by language models, TOUCAN and APIBench. On TOUCAN at privacy budget $\varepsilon=4$ with the Jina encoder, Hit@5 (the share of queries with a correct tool in the top five) rises from 32.3\% with documentation alone to 57.2\% on labeled test queries, whose correct tools have usage records. The gain is smaller with a second encoder and on APIBench. On labeled queries in both catalogs, the privately trained pipeline beats a public generator given the same number of requests per tool. On TOUCAN labeled queries, \method{} also beats public documentation expansion, in which an untrained model writes the same number of requests for every tool (one generated index of similar size); on APIBench this gain is detected with one encoder only. On TOUCAN held-out queries, public expansion is better with Jina. Also on TOUCAN, rules that add the noisy counts to the score, at their development-selected weights, raise labeled retrieval but lower retrieval for tools without records more than \method{} does. In post hoc tests, adding the counts to the public generator's requests beats \method{} on labeled queries. Adding them to \method{} instead keeps it ahead of that rule on tools without records; on labeled queries it trails that rule by an amount whose interval includes zero with Jina and by about four points with Nomic. At $\varepsilon=4$ on TOUCAN's primary held-out queries, tools without records lose about two points; intervals clustered by query detect this loss, intervals clustered by tool do not, and neither bounds it within two points. On APIBench they gain about three points over documentation, not detectably more than with public expansion.
\end{abstract}


\section{Introduction}\label{sec:intro}
An agent that can call thousands of tools cannot read every tool's documentation for every request. It first retrieves a handful of candidate tools and chooses among those. Dense retrievers built for ordinary search do poorly at this step \citep{shi2025toolret}. Example requests for each tool help, whether written from documentation \citep{chen2024reinvoke} or taken from usage \citep{moon2024tool2vec,anon2026zoe}. Requests users actually sent are a natural source of such examples. They are often private, and they sit with the many organizations that run the agents.

Consider several clinics whose agents share one public catalog of tools. A lab-results lookup is used every day, and an allergy-registry tool almost never. The clinics would all retrieve better if the index knew how the lab-results tool is asked for. They cannot pool their request logs, because each request can carry patient details. They can share the catalog's public documentation, and they could share outputs protected by differential privacy (DP), which bounds how much any single record can change what is released \citep{dwork2014algorithmic}.

Two simple ways to share usage under DP leave gaps. Noisy per-tool counts are cheap to release, but they carry mostly popularity. In our measurements, a rule that ranks by the released counts helps frequently used tools but, at the weight chosen on development data, lowers retrieval for queries about tools without records by about eleven points (twelve with a full-budget release) on the primary held-out queries of our second TOUCAN test set (Section~\ref{sec:counts}). DP-trained language models can write synthetic queries that keep a formal guarantee \citep{yue2023dpsynthetic}, and \citet{carranza2024dpqueries} train a retriever on such queries for one data holder. That work does not build a shared tool index from several organizations or ask what happens to tools without usage records.

\method{} combines the two releases in one index. The clients jointly fine-tune a request generator with differentially private stochastic gradient descent (DP-SGD) and release noisy per-tool usage counts. Under secure aggregation, a protocol that reveals only the sum of client inputs and which we simulate, only noised sums leave the clients. The counts decide how many requests the generator writes for each tool, and those generated requests become the tool's usage field. A per-field rule scores the usage field beside the tool's public documentation, and a tool without generated usage keeps its documentation score. Allocation, generation, indexing and scoring use only released outputs and public documentation, so for fixed scoring settings and secret noise the guarantee covers the whole index. Our benchmark uses public noise seeds, so it measures utility only, and the reported $\varepsilon$ covers one release (Section~\ref{sec:method}).

We ask four questions, using TOUCAN \citep{xu2026toucan}, a catalog of 2,159 Model Context Protocol tools, two embedding models (Jina and Nomic) and four privacy budgets. TOUCAN's requests, like those of our second catalog, were written by language models, so our experiments test how the mechanism behaves, not the value of real users' requests.

RQ1: Does \method{} help retrieval for tools with usage records? Yes. We measure Hit@5, the share of queries with a correct tool among the five top-ranked tools, on labeled queries, whose gold tools have records, and on held-out queries, whose gold tools have none. At $\varepsilon=4$ with the Jina embedding model, Hit@5 on labeled test queries rises from 32.3\% with documentation alone to 57.2\%. The lower bound of the gain's interval, Bonferroni-adjusted for the eight encoder and budget settings, is above zero in all eight. A second test set of fresh queries shows the gain again (Section~\ref{sec:rq1}).

RQ2: What does private usage add over public generation? A public generator given the same number of requests per tool has lower labeled Hit@5 than the privately trained pipeline in all eight settings, by 3.6 to 16.2 points in point estimate. On the second TOUCAN test set, \method{} also beats public documentation expansion, in which an untrained generator writes the same number of requests for every tool, and a variant of Re-Invoke \citep{chen2024reinvoke} in which a 7B model writes ten requests per tool, on labeled queries with both encoders. With Jina, public expansion retrieves tools without records better, by up to 8.9 points; with Nomic we detect no difference (Section~\ref{sec:rq2}).

RQ3: Does \method{} do better than counts released at the same budget? At the weight chosen on development data, a count-only rule on \method{}'s own count release comes within a few points of the labeled gain of \method{} on Jina and exceeds it on Nomic, but it ranks tools without records lower. A post hoc sweep of its weight finds no setting ahead of \method{} on both kinds of queries. In post hoc tests, adding the counts to the public generator's requests raises labeled Hit@5 above \method{} but is about eight held-out points below it. Adding them to \method{} instead stays about six held-out points above that baseline, though still below documentation, and trails it on labeled queries (Section~\ref{sec:counts}).

RQ4: What does \method{} cost tools without records? At $\varepsilon=4$, retrieval for these tools falls about two points below documentation in point estimate, on both encoders and both test sets. On all queries about tools without records, the drop is about one point. On the primary held-out population, intervals clustered by query detect this loss on both encoders; intervals clustered by tool include zero, and neither bounds it within two points (Section~\ref{sec:rq3}).

We repeated the study at $\varepsilon=4$ on a second catalog, APIBench \citep{patil2023gorilla}, with its settings and comparisons fixed before any run. The labeled gains over documentation and over the public generator hold on both encoders, at a smaller size. The labeled gain over public expansion is detected with Nomic only, so the planned two-encoder criterion is not met. Held-out Hit@5 rises by about three points there (Section~\ref{sec:apibench}).

Our contributions are \method{}, a design for sharing tool-use experience across organizations whose mechanism has a record-level DP guarantee (Section~\ref{sec:method}), and evidence that it raises labeled retrieval on two catalogs and, on TOUCAN, loses less on tools without records than count rules at their development-selected weights, with every planned and post hoc comparison labelled as such (Sections~\ref{sec:rq1} to~\ref{sec:apibench}).

\section{Related work}\label{sec:related}

\paragraph{Tool retrieval and usage-aware tool representations.}
Two lines of work add text about how a tool is used: queries generated from documentation \citep{chen2024reinvoke} and example user queries \citep{moon2024tool2vec} (Table~\ref{tab:positioning}). Both are related to document expansion in ad hoc search, where generated queries are appended to a document before indexing \citep{gospodinov2023doc2queryminus}. Earlier work found that folding usage examples into a tool's text lowers retrieval for tools that have no usage, and that scoring documentation and usage as separate fields reduces that loss \citep{anon2026zoe}. None of this work studies how to combine usage records held by several organizations under a privacy guarantee.

\paragraph{Sharing agent knowledge across organizations.}
Several systems already move agent experience between organizations or agents (Table~\ref{tab:positioning}). FICAL clients exchange agent-written knowledge compendiums; its authors argue that these leave out private user information but give no formal bound \citep{wu2024fical}. ToolAtlas fills a tool-side memory by having the provider probe its own tools \citep{fang2026toolatlas}, and federated routers share model updates or cluster statistics without a formal privacy analysis \citep{askin2026federaterouter}. None of these systems shares user requests under a formal privacy guarantee.

\paragraph{Differential privacy for retrieval and synthetic text.}
Fine-tuning a pretrained language model with differentially private stochastic gradient descent (DP-SGD; \citealp{abadi2016dpsgd}) keeps much of its non-private utility, especially with parameter-efficient updates \citep{yu2022dpfinetune,li2022dplearners}. Sampling from such a model gives synthetic text that carries the same guarantee \citep{yue2023dpsynthetic}. \citet{carranza2024dpqueries} use this recipe to generate queries from documents and train a deep retriever on them, and \citet{koga2024dprag} protect the retrieval corpus at answer time in retrieval-augmented generation. Whether private synthetic usage helps tools that have usage without pushing down tools that have none is the open question we test.

\paragraph{Closest work.}
Table~\ref{tab:positioning} compares \method{} with the closest methods, as we read each paper. The closest, \citet{carranza2024dpqueries}, shares our generator recipe: a DP-SGD model writes queries from documents, which train a retriever for one data holder. Our queries become index entries for a catalog shared by several organizations, a second private release (noisy counts) decides how many each tool gets, and we report tools with and without records separately. Our experiments vary the usage source within one fixed scoring pipeline and make no claim against complete Carranza-style retriever training, Tool2Vec or full Re-Invoke, which add a trained retriever or a query-side component.

\begin{table}[t]
\centering\footnotesize
\setlength{\tabcolsep}{4pt}
\begin{tabular}{lllll}
\toprule
Method & Text about each tool & Across orgs. & Formal privacy & Used as \\
\midrule
Re-Invoke \citep{chen2024reinvoke} & Queries from documentation & No & Not needed & Index entries \\
Tool2Vec \citep{moon2024tool2vec} & Example user queries & No & No & Tool embeddings \\
Per-field scoring \citep{anon2026zoe} & Usage examples & No & No & Separate field \\
\citet{carranza2024dpqueries} & DP queries from documents & One holder & DP & Retriever training \\
FICAL \citep{wu2024fical} & Agent-written compendiums & Yes & No formal bound & Shared knowledge \\
ToolAtlas \citep{fang2026toolatlas} & Provider probes & Across agents & Not applicable & Tool-side memory \\
Fed.\ routers \citep{askin2026federaterouter} & Client evaluations & Yes & No formal analysis & Router training \\
\method{} (ours) & DP requests, DP counts & Simulated & Record-level DP$^*$ & Separate field \\
\bottomrule
\end{tabular}
\caption{Among the listed methods, only \method{} shares usage across organizations under a formal guarantee and indexes it as a separate field. Each cell is our reading of the cited paper. orgs.: organizations. Fed.\ routers: federated routers that choose among language models. Not needed: Re-Invoke uses only public documentation. Not applicable: ToolAtlas shares no user requests. Simulated: five hash partitions of one pool. $^*$: under the assumptions of Appendix~\ref{app:noise}; our benchmark releases use public noise seeds and carry no guarantee. Only Re-Invoke is run, as a restricted variant without its query-side intent extractor; Tool2Vec's learned tool embeddings and Carranza-style retriever training are not run, and the per-field rule is reused.}
\label{tab:positioning}
\end{table}

\paragraph{Distributed differential privacy.}
Secure aggregation lets a server learn only the sum of client vectors \citep{bonawitz2017secagg}. When each client adds a share of the noise before that sum, the server sees a noised total, and accounting follows the central model approximately \citep{mcmahan2018dpfedavg}; a deployment needs a discrete mechanism, whose accounting differs from the continuous Gaussian we simulate \citep{kairouz2021ddg}.

\section{\method}\label{sec:method}

\begin{figure}[t]
\centering
\includegraphics[width=0.88\linewidth]{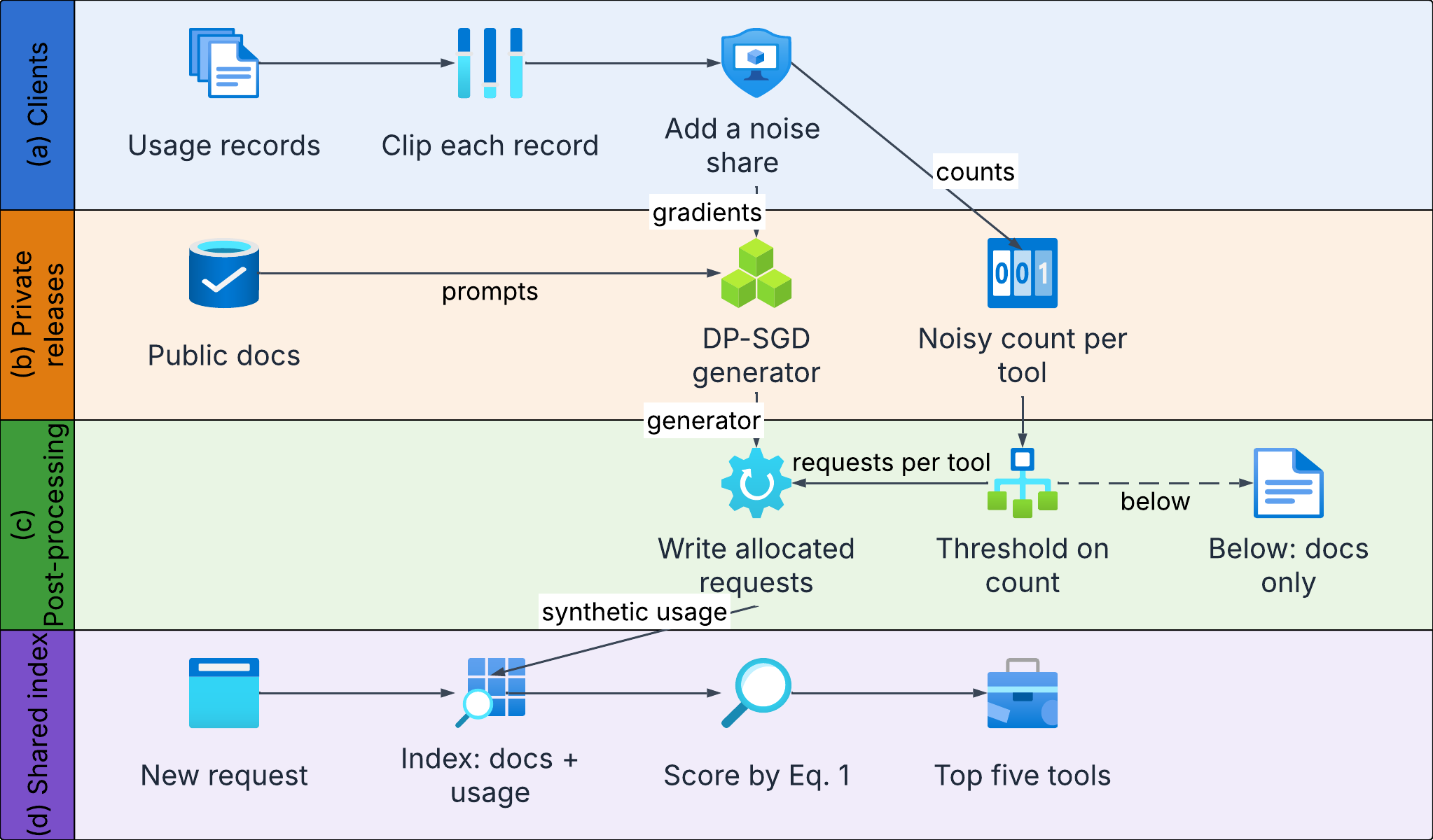}
\caption{Private synthetic usage pipeline. (a) Clients clip each usage record and add a share of the noise. (b) Secure aggregation, simulated here by exact summation, releases a DP-SGD generator and a noisy count per tool, splitting $\varepsilon$ evenly. (c) A threshold on the noisy count sets how many synthetic requests $n_t$ each tool receives; tools below it get none. (d) The index scores a new request with Eq.~\ref{eq:score} and returns the top five tools; tools without synthetic usage keep their documentation score, although their ranks may change. Rows (c) and (d) use only released outputs and public documentation (Algorithm~\ref{alg:synapse}).}
\label{fig:workflow}
\end{figure}

\method{} gives each tool in a shared index two fields: the tool's public documentation, and usage entries. Under privacy, the usage entries are synthetic requests written by a language model trained with differential privacy on the clients' requests, and a per-field rule scores the two fields separately. The private pipeline is new in this paper. The scoring rule comes from earlier workshop work \citep{anon2026zoe}. In Figure~\ref{fig:workflow}, clients clip and noise their records (a), two private releases leave them (b), post-processing allocates and generates requests (c), and the shared index scores new requests (d).

\subsection{Setting and running example}\label{sec:setting}

Five clinics run agents over a shared registry of tools. Each tool has public documentation, for example a lab-results lookup whose description says what it returns. Each clinic also logs usage records. A usage record is one user request together with the tools that served it, its gold tools. A request such as ``latest potassium panel for the patient seen on Tuesday'' says something about one patient. The clinics want a shared tool index in which each tool carries usage text, because usage text helps match new requests to tools. They want a formal bound on how much adding or removing one request can change the released output. This is not a guarantee that no request can ever be inferred or reproduced.

Formally, a registry $\mathcal{T}$ of $N$ tools gives each tool $t$ public documentation $d_t$. Client $k \in \{1,\dots,K\}$, with $K=5$, holds a set $R_k$ of usage records, and $R=\bigcup_k R_k$. A record $r=(x_r,G_r)$ has request text $x_r$ and gold tools $G_r \subseteq \mathcal{T}$. We keep at most eight gold tools per record.

\paragraph{Privacy unit and trust model.}
Two datasets are neighbors if one equals the other with a single record added or removed, at any client. All guarantees in this paper are $(\varepsilon,\delta)$-differential privacy for this record-level unit, with $\delta=10^{-6}$ in total. A record can name several tools, so the guarantee covers a request and all of its gold tools together. It does not cover a user who sends many requests. No server is trusted with a single client's contribution. Each client adds a share of the noise to what it sends, and secure aggregation reveals only the sum \citep{bonawitz2017secagg,kairouz2021ddg}. Shares are calibrated so that any three honest clients out of five supply the full noise. With five honest clients the noise is $\sqrt{5/3}$ times the target, and our results use this actual noise. Our experiments run all clients in one process and simulate secure aggregation as an exact sum, so this trust model is assumed, not enforced by our code; the simulator also logs each step's realized sample count, which our accounting does not cover and a deployment must not reveal. The guarantee is for one release at fixed scoring settings: the usage weight $w$ (Section~\ref{sec:scoring}), the threshold $\tau$ (Section~\ref{sec:allocation}) and the count weight $\lambda$ of the count-only rule (Section~\ref{sec:setup}). Every $\varepsilon$ we report is this per-release budget. We selected the settings with a separate calibration release trained on the same pool, so a deployment that calibrates and then publishes one release on the same records spends $(2\varepsilon, 2\delta)$ under basic composition, if its development data are public or fixed independently of the pool (Appendix~\ref{app:noise}). The guarantee covers whether a record is in the pool. It does not hide a clinic's overall mix of cases, which the released totals still reflect. It also assumes secret noise. Our benchmark draws noise from public seeds, so anyone can subtract it from a released count; our releases simulate the mechanism's utility and carry no guarantee. Publishing several releases from the same records composes their budgets; Appendix~\ref{app:noise} states the guarantee and its scope formally.

\subsection{Per-field scoring}\label{sec:scoring}

The index keeps documentation and usage as separate fields of each tool and scores them separately. We use the per-field rule of earlier workshop work without change \citep{anon2026zoe}. For a query $q$ and an encoder $f$, the documentation score is $a_t(q)=\cos(f(q),f(d_t))$. If tool $t$ has a set $U_t$ of usage entries, its usage score is the best match, $b_t(q)=\max_{u\in U_t}\cos(f(q),f(u))$. For each query we standardize the documentation score over all tools, giving $z^{\mathrm{doc}}_t(q)$, and the usage score over the tools that have usage entries, giving $z^{\mathrm{use}}_t(q)$. Tools without usage are left out of the second standardization rather than counted as zero. Both use the population standard deviation, replaced by one when it is zero. The final score is
\begin{equation}
s_t(q) = (1-w)\, z^{\mathrm{doc}}_t(q) + w\, z^{\mathrm{use}}_t(q)\quad\text{if } U_t\neq\emptyset,\qquad s_t(q)=z^{\mathrm{doc}}_t(q)\quad\text{otherwise},
\label{eq:score}
\end{equation}
with one weight $w\in[0,1]$ tuned on development data.
A tool without generated usage keeps its documentation score. Its rank can still fall when tools with generated usage move above it, and our measurements report that change.

The rule needs individual usage entries, because it scores the single best match. A per-tool sum of usage embeddings would be cheap to release privately, since one record changes it by a bounded amount. On our development data such sums behave mostly like a popularity count; tuned together with counts, the summed usage embedding got weight zero (Appendix~\ref{app:dev}, Table~\ref{tab:synapse-aggregate-dev}). We therefore keep the per-field rule and give it private synthetic usage entries instead of real ones.

\subsection{Private synthetic usage}\label{sec:private}\label{sec:allocation}

One release consists of noisy per-tool usage counts and a DP-trained request generator. The total budget $\varepsilon$ is split evenly between these two mechanisms.

\paragraph{Count release.}
Each record contributes a vector $v_r\in\mathbb{R}^N$ with $(v_r)_t=|G_r|^{-1/2}$ for $t$ in its set $G_r$ of at most eight retained gold tools and zero elsewhere, so $\|v_r\|_2=1$. A record with one gold tool adds one unit to that tool. A record with four gold tools adds one half to each. The clients release the noisy sum $\tilde c=\sum_{r\in R} v_r + \mathcal{N}(0,\sigma_c^2 I)$ for every tool in the public registry, including tools with no records. Adding or removing one record changes the sum by at most one in $\ell_2$ norm. The analytic Gaussian mechanism \citep{balle2018analytic} calibrates a central target $\sigma_{c,\mathrm{target}}$ at $(\varepsilon/2,\,10^{-7})$. When all five clients are honest, the sum has $\sigma_c=\sqrt{5/3}\,\sigma_{c,\mathrm{target}}$, and we report results at this noise.

\paragraph{Generator.}
We fine-tune a public instruction model with LoRA adapters under DP-SGD \citep{abadi2016dpsgd,yu2022dpfinetune}. Each training example maps the public documentation of a gold tool to the request, so a record with $g$ gold tools gives $g$ examples. The per-record gradient is the sum over those $g$ examples, clipped to norm one as a whole. This keeps one record's influence on each step bounded no matter how many tools it names. At each of a fixed number of steps, every client includes each of its records independently with a fixed probability $q$, sums the clipped gradients of the included records and adds its noise share, even when it includes none; the optimizer receives the aggregate divided by the fixed expected batch size. The probability, the step count and the divisor are constants set before training from the public pool size, so they do not depend on which records are present. A numerical accountant \citep{gopi2021prv} sets the noise multiplier from its upper bound for the generator's half of $\varepsilon$ and $\delta=9\times10^{-7}$. A R\'enyi accountant serves only as a cross-check. The conditioning prompt contains only public documentation.

\paragraph{Allocation and generation.}
Let $\sigma$ be the noise level actually in $\tilde c$. Tool $t$ receives $n_t=\lfloor\max(\tilde c_t,0)+1/2\rfloor$ synthetic requests if $\tilde c_t \ge \tau\sigma$ and none otherwise, with the threshold $\tau$ tuned on development data. The generator samples $n_t$ requests conditioned on $d_t$, and these become the usage entries $U_t$ in Eq.~\ref{eq:score}. Allocation, generation, indexing and scoring use only released outputs and public documentation, so they are post-processing and add no privacy cost. The threshold is not needed for privacy. It exists because noise alone gives some tools with no records a positive count. At the thresholds selected for $\varepsilon=4$ ($\tau=1$ on Jina, $\tau=0$ on Nomic), 257 to 266 and 686 to 703 of the 1,602 TOUCAN tools without records receive synthetic usage they never had (Appendix~\ref{app:noise}, Table~\ref{tab:first-test-allocation}).

Algorithm~\ref{alg:synapse} lists the steps: the count release above, DP-SGD training of the generator ($\theta$ denotes its adapter parameters), sampling at temperature 0.7 from the public documentation alone, and the per-field rule.

\begin{algorithm}[t]
\caption{Building the \method{} index for one release. Only lines 1 to 3 touch private records.}
\label{alg:synapse}
\begin{algorithmic}[1]
\Require client records $R_1,\dots,R_K$; documentation $\{d_t\}$; budget $(\varepsilon,\delta)$; weight $w$; threshold $\tau$
\State $\tilde c \gets \textsc{CountRelease}(R, \varepsilon/2, 10^{-7})$ \Comment{noisy per-tool counts}
\State $\theta \gets \textsc{TrainGenerator}(R, \varepsilon/2, 9\times10^{-7})$ \Comment{DP-SGD with LoRA}
\State release $(\tilde c, \theta)$; $\sigma \gets$ noise level in $\tilde c$
\For{each tool $t$}
  \State $n_t \gets \lfloor \max(\tilde c_t,0)+1/2 \rfloor$ if $\tilde c_t \ge \tau\sigma$, else $0$ \Comment{allocation}
  \State $U_t \gets n_t$ samples from $\textsc{Generate}(\theta, d_t)$
\EndFor
\State index each tool by $f(d_t)$ and $\{f(u) : u \in U_t\}$
\State \Return $\textsc{Score}(q)$: Eq.~\ref{eq:score} with weight $w$, top five tools
\end{algorithmic}
\end{algorithm}

In the running example, the lab-results lookup has many records, so its noisy count is large and the generator writes requests for it from its documentation. A new request matches the closest of those synthetic requests. A rarely used allergy-registry tool may fall below the threshold and keep its documentation score alone. No raw clinic record enters the shared index. The DP-trained model writes the lab-results tool's usage text. DP does not rule out verbatim overlap with a training record, so we report overlap and do not filter generated requests by it.

\section{Experimental setup}\label{sec:setup}

\paragraph{Catalogs and query strata.}
TOUCAN is our primary catalog \citep{xu2026toucan}. Its tools come from real Model Context Protocol (MCP) servers, and its tasks were written by language models. Section~\ref{sec:apibench} repeats the study on a second catalog. Our usage pool holds 4,306 records over 2,159 tools (construction rules in Appendix~\ref{app:data}). Of these tools, 1,602 have no records, and the most-used tool has 251. We evaluate on two query strata. Held-out queries have gold tools with no records in the pool. They measure what sharing does to tools nobody has used. Labeled queries have gold tools with records. Development sets have 1,207 held-out and 2,879 labeled queries. The first test set has 769 held-out and 2,595 labeled queries. Of the labeled test queries, 1,587 have at least 20 records for every gold tool. The five clients are simulated, not observed organizations. A fixed hash of each record's identifier assigns the whole record, with all of its gold tools, to one client, which gives TOUCAN clients between 831 and 898 records.

\paragraph{Retrieval and metric.}
We embed queries, documentation and usage with Jina Embeddings v2 \citep{gunther2023jina} and Nomic Embed \citep{nussbaum2025nomic}. Hit@5 is the share of queries with at least one gold tool among the five highest-scoring distinct tools.

\paragraph{Arms.}
All arms share the encoder, registry and fixed query splits. The per-field arms use Eq.~\ref{eq:score}. Development-only aggregate controls built from the pool records keep their own fixed scoring rules (Section~\ref{sec:counts}, Appendix~\ref{app:dev}).
Documentation only sets $w=0$ and is the reference on both strata. \method{} applies the per-field rule to private synthetic usage (Section~\ref{sec:private}). The public generator uses the same base model without private fine-tuning, the same prompt template and the same noisy allocation, with its own $w$ tuned on development data; its difference from \method{} compares two tuned pipelines that differ in private training, and because it uses the private count release it is not free of privacy cost. Public expansion (an untrained generator, equal requests per tool) and a Re-Invoke-style variant (Mistral-7B \citep{jiang2023mistral}, ten requests per tool, averaged expanded-document embeddings) use no private data (Section~\ref{sec:rq2}). The count-only rule, with half or full budget, adds $\lambda\log(1+\max(\tilde c_t,0))$ to $z^{\mathrm{doc}}_t(q)$ and uses no generated text (Section~\ref{sec:counts}).

\paragraph{Generator and training.}
The primary generator is Llama-3.2-1B-Instruct, from the Llama 3 family \citep{grattafiori2024llama3}, with LoRA of rank 4 on the query and value projections. AdamW uses learning rate $4\times10^{-4}$ for 673 fixed Poisson steps (nominally 20 epochs), with sampling probability $128/4306$ and expected batch 128. The joint record clip is one, the context limit is 512 tokens, and the target limit is 128 tokens. Label smoothing is 0.1 (Appendix~\ref{app:training}). We decode at temperature 0.7. Table~\ref{tab:noise} gives the noise we add. We fixed all hyperparameters before any run on the usage pool and did not tune them on it, since every tuning run on private data would spend budget.

\paragraph{Tuning and runs.}
The weight $w$ ranges over $\{0,0.1,\dots,1\}$ and the threshold $\tau$ over $\{0,1,\dots,5\}$, in units of the noise actually in the count release. We pick the pair that maximizes the average of labeled and held-out development Hit@5, each weighted one half. Ties go to the smaller $w$, then the larger $\tau$. For each $\varepsilon$, one calibration release trains, releases and generates, and we tune $w$ and $\tau$ on development queries with it alone. Three further evaluation releases are scored on test with those settings. Appendix~\ref{app:impl} lists the prompts, every arm's grid and the selected settings.

\paragraph{Test sets and statistics.}
Each of the first test set's eight settings (two encoders by four budgets) uses three releases. A second test set uses queries that no earlier run had scored and six releases at $\varepsilon=4$, three of them shared with the first test. Its primary held-out population leaves out every query with a gold tool that is also gold in development, which gives 2,237 queries over 247 tool clusters. A second held-out population, all zero-support, keeps all 6,097 queries whose gold tools have no records. The second test set also has 15,891 labeled queries. All settings were fixed on development data before their test scoring. Seven analyses came after test results and are labelled post hoc: on TOUCAN, full-budget counts and the count sweep (both planned in advance on APIBench); the public generator with counts, \method{} with counts, a near-copy check, a combined index and an APIBench rank diagnostic. All inference holds the pool, client partition, model initialization and calibration outcome fixed. Intervals resample whole releases and, independently in each bootstrap draw, either query identities or tool clusters, which group queries by their lexicographically smallest gold tool. We show tool-clustered intervals. First-test intervals are simultaneous over the eight settings by a Bonferroni adjustment. Second-test intervals on the primary population are simultaneous over the two encoders. All-zero-support intervals are descriptive 95\% intervals. We describe the change on held-out queries against a two-point margin fixed in advance.

\section{RQ1: \method{} helps retrieval for tools with records}\label{sec:rq1}
\begin{figure}[t]
\centering
\includegraphics[width=\linewidth]{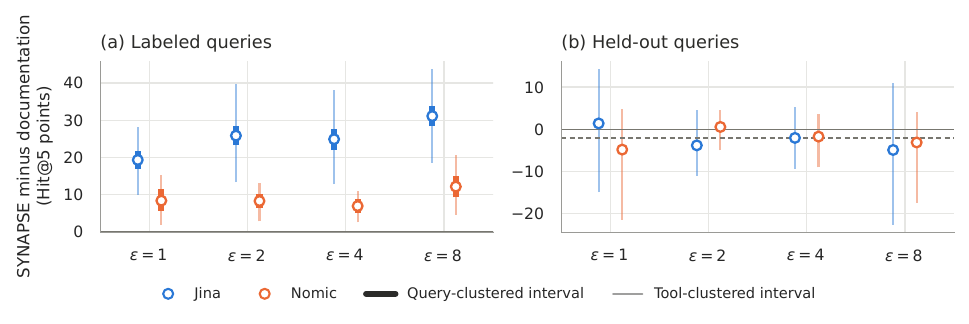}
\caption{\method{}'s labeled gain over documentation is above zero in all eight first-test settings, and every held-out interval includes both zero and the $-2$-point margin (dashed). \method{} minus documentation, Hit@5 points, mean of three releases per setting, with intervals simultaneous over the eight settings: query-clustered (thick) and tool-clustered (thin). Values: Table~\ref{tab:first-test-gate}.}
\label{fig:first-test}
\end{figure}

\method{} raised labeled Hit@5 above documentation in all eight first-test settings (Figure~\ref{fig:first-test}). The simultaneous lower bound of the gain is above zero in every setting, under both query and tool clustering (Appendix~\ref{app:labeled-ci}, Table~\ref{tab:first-test-labeled-ci}). At Jina $\varepsilon=4$, labeled Hit@5 rises from 32.29\% to 57.20\% (Table~\ref{tab:first-test-gate}).

The gain is largest, descriptively, on tools with at least 20 records; on the rest, three of four Nomic intervals include zero (Appendix~\ref{app:labeled-ci}).

On the second test set (six $\varepsilon=4$ releases), \method{} scores 57.35\% against 32.50\% for documentation on Jina and 60.23\% against 50.10\% on Nomic. A planned non-private reference, which indexes the real pool requests with the same rule at its development weight, scores 58.89\% and 57.85\% (Table~\ref{tab:fresh-test}).

\begin{table}[t]
\centering\small
\setlength{\tabcolsep}{4pt}
\begin{tabular}{lrrrrrr}
\toprule
 & \multicolumn{3}{c}{Jina} & \multicolumn{3}{c}{Nomic} \\
\cmidrule(lr){2-4}\cmidrule(lr){5-7}
Arm & Labeled & Held-out & Zero-supp. & Labeled & Held-out & Zero-supp. \\
\midrule
Documentation & 32.50 & 46.89 & 56.60 & 50.10 & 65.40 & 69.90 \\
Public expansion & 41.22 & \underline{53.15} & \underline{63.13} & 51.73 & \underline{65.49} & 71.51 \\
Re-Invoke-style & 36.67 & 51.23 & 61.96 & 53.04 & 64.73 & \underline{73.00} \\
\midrule
Public gen., private alloc.\ & 47.19 & 44.23 & 54.89 & 56.37 & 60.01 & 67.11 \\
Count (half) & 53.09 & 35.54 & 44.02 & 66.00 & 54.31 & 59.84 \\
Count (full)$^\dagger$ & 54.08 & 34.43 & 43.68 & 66.72 & 53.32 & 59.41 \\
Public gen.\ + counts$^\dagger$ & \underline{65.17} & 35.70 & 44.46 & \underline{70.85} & 55.20 & 60.98 \\
\midrule
\method{} & 57.35 & 44.28 & 55.45 & 60.23 & 63.25 & 69.05 \\
\method{} + counts$^\dagger$ & 64.42 & 42.34 & 53.44 & 66.72 & 61.33 & 66.73 \\
\midrule
$\Delta$ \method{} $-$ doc. & $+24.84$ & $-2.62$ & $-1.15$ & $+10.13$ & $-2.15$ & $-0.85$ \\
Non-private reference & 58.89 & 48.95 & 60.65 & 57.85 & 69.65 & 75.45 \\
\bottomrule
\end{tabular}

\caption{On the TOUCAN second test set, no arm has the highest Hit@5 on both labeled and held-out queries. Hit@5 (\%), $\varepsilon=4$, mean of six releases; row groups use no private data, private counts only, and private text and counts; underlined: column maximum, reference excluded; $^\dagger$: post hoc. Public gen., private alloc.: the untrained base generator with \method{}'s noisy allocation. Count (half), Count (full): count-only rule at $\lambda=0.3$, half or full budget. Queries: 15,891 labeled, 2,237 primary held-out (Section~\ref{sec:setup}), 6,097 zero-support. $\Delta$: from unrounded values. Reference: real pool requests, planned, no intervals. Public arms were generated once.}
\label{tab:fresh-test}
\end{table}

\section{RQ2: what private usage adds over public generation}\label{sec:rq2}
Private-minus-public labeled Hit@5 is positive in all eight first-test settings, by 3.58 to 16.16 points (Appendix~\ref{app:private-public}). Each setting's descriptive 95\% tool-clustered interval excludes zero; we report them one by one, not as a second eight-comparison family. On held-out queries the private-minus-public point estimate is negative in five of eight settings.

The second test set shows the same labeled ordering on both encoders (Table~\ref{tab:fresh-test}). The comparison fixes the prompt template, base model, noisy counts and allocation; rendered date lines may differ in two first-test releases (Appendix~\ref{app:data}). The retrieval weight $w$ was tuned separately for each generator on development data. On Nomic both selected $w=0.5$ and $\tau=0$, so there the comparison is at a fixed weight. On Jina the weights differ (0.8 and 0.9), and the comparison is between two tuned pipelines. In a post hoc check, removing the 30 of 2,595 first-test labeled queries whose Jina cosine to any generated request is at least 0.95 leaves the private-minus-documentation and private-minus-public gains positive in all eight settings, with descriptive tool-clustered intervals excluding zero; broader effects of the benchmark's writing style are not tested.

\paragraph{Public documentation expansion is the closest practical alternative.}
An organization that cannot share usage can still expand each tool's documentation with requests that a public model writes from that documentation \citep{chen2024reinvoke}. If such public expansion reached \method{}'s Hit@5 for a given index, encoder and query population, private usage would add no detected gain there. We therefore fixed two public-expansion arms in a protocol before generating any request.
Public expansion uses \method{}'s base generator without fine-tuning, with the same prompt and decoding, and writes 3 requests per tool on TOUCAN (11.2\% fewer in total than \method{}) and 9 on APIBench (1.4\% fewer), counts set from \method{}'s mean index size over releases that already existed; every tool then has usage, and $w$ is tuned on development data over 11 weights, against 66 settings for \method{} (Table~\ref{tab:design}). The Re-Invoke-style arm follows the query generator of \citet{chen2024reinvoke} but omits its query-side intent extractor, so it is a variant, not a reproduction.
Neither arm uses private data or privacy budget. Each public index was generated once, so intervals hold it fixed and omit generation variability. Tables~\ref{tab:fresh-test} and~\ref{tab:apibench} report both arms next to documentation, \method{} and the public generator of the previous paragraphs, which uses the private allocation; Table~\ref{tab:public-expansion-ci} in Appendix~\ref{app:public-expansion} gives the differences with intervals.

\paragraph{On TOUCAN, \method{} beats public expansion on labeled queries.}
\method{} exceeds public expansion by 16.12 points on Jina and 8.50 on Nomic, and the Re-Invoke-style arm by 20.68 and 7.19 points. All four intervals exclude zero, each simultaneous over the two encoders for its own comparator, and the narrowest lower bound is 1.86 points. The public index is 11.2\% smaller; we did not test a matched index.

\paragraph{With Jina, public expansion retrieves tools without records better.}
With Jina, public expansion exceeds \method{} by 8.87 points on primary held-out queries, interval $[-14.89, -3.11]$ for \method{} minus public expansion, and by 7.68 points on all zero-support queries, $[-11.94, -3.66]$. The Re-Invoke-style arm is ahead by 6.51 points on all zero-support queries, $[-13.05, -0.27]$. With Nomic, both public arms are ahead by 1.5 to 4.0 points, and every interval includes zero.

A post hoc combined index gives tools below the count threshold public requests instead. With Re-Invoke-style requests it exceeds \method{} on TOUCAN labeled queries by 9.89 points on Jina, $[7.36, 12.52]$, and 5.57 on Nomic, $[3.34, 8.49]$, with primary held-out intervals that include zero; its indexes are larger, 10,346 to 22,204 requests against about 7,300 (Appendix~\ref{app:public-expansion}).

Both catalogs' requests were written by language models \citep{xu2026toucan}, so these comparisons do not establish the value of knowledge that only real users' requests contain.

\section{RQ3: \method{} against a count-only rule}\label{sec:counts}
Counts are the cheapest private signal of usage, so we ask whether \method{} does better than counts alone. The count-only rule adds $\lambda\log(1+\max(\tilde c_t,0))$ to the documentation score and uses no generated text. We test it on two count releases. The half-budget rule reuses \method{}'s count release and was registered, with its weight, before the second test set was scored. The full-budget rule spends the whole $\varepsilon=4$ on a separate count release, the same total budget as \method{}, and is post hoc. For each, development data select $\lambda=0.3$ from one calibration release, and six evaluation releases are scored on the second test set.

\paragraph{At the development-selected weight, counts trade held-out tools for popular ones.}
Against the registered half-budget rule, \method{} differs on labeled queries by $+4.26$ points on Jina, descriptive 95\% interval $[-0.78, 9.30]$, and by $-5.77$ on Nomic, $[-9.50, -2.64]$. The post hoc full-budget rule gives similar differences (Appendix~\ref{app:count-sweep}).

The held-out queries reverse the picture. On the primary held-out population, \method{} is above the half-budget rule by 8.74 points on Jina, $[4.84, 13.79]$, and 8.95 on Nomic, $[4.78, 14.89]$. Every interval on all zero-support queries also excludes zero.

\begin{figure}[!tb]
\centering
\includegraphics[width=\linewidth]{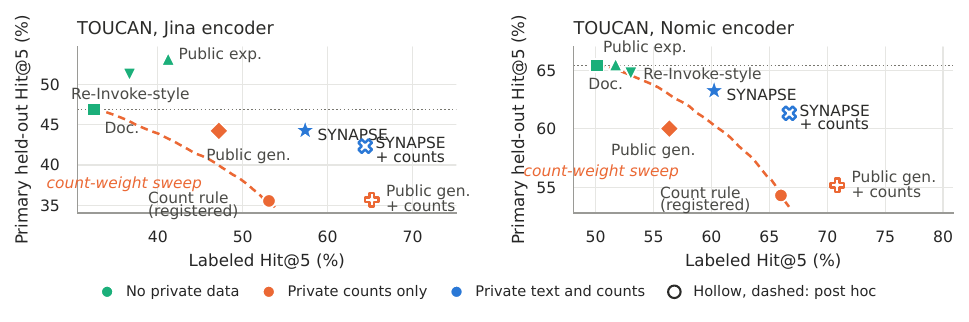}
\caption{No arm has the highest Hit@5 on both labeled and primary held-out TOUCAN queries. Second test set, $\varepsilon=4$, point estimates from Table~\ref{tab:fresh-test}; intervals for the differences are in Figure~\ref{fig:second-test-forest}. Colour: private data used. Hollow markers and dashed curve: post hoc. Dotted line: documentation's held-out Hit@5. The curve varies the count weight $\lambda$ from 0 (documentation) to 0.3 on the full-budget release.}
\label{fig:tradeoff}
\end{figure}

\paragraph{In a post hoc sweep, no count weight reaches \method{} on both kinds of queries in point estimate.}
After the second test set had been scored, we swept $\lambda$ from 0 to 0.3 on the full-budget release (Figure~\ref{fig:tradeoff}, Appendix~\ref{app:count-sweep}). On both encoders, every tested weight is below \method{} on labeled or on held-out queries in point estimate. The intervals are pointwise and the sweep covers one popularity rule, so we read it as a diagnostic, not a planned test.

\paragraph{The public generator with a count prior gains labeled and loses held-out retrieval.}
After the test results, we ran a post hoc baseline that adds the half-budget counts to the score of the public generator, whose requests the same counts already allocate. Its settings were tuned on development data over the same $w$ and $\lambda$ grids as below plus six thresholds, so its search is larger than that of \method{} with counts, which keeps \method{}'s threshold (Appendix~\ref{app:impl}). Against \method{} it gains 7.82 labeled points on Jina, descriptive 95\% tool-clustered interval $[4.49, 11.19]$, and 10.62 on Nomic, $[6.66, 14.71]$. It loses 8.58 primary held-out points, $[-13.08, -5.17]$, and 8.05, $[-13.08, -4.00]$.

\paragraph{With the same count prior, \method{} keeps its held-out lead.}
We then wrote down and ran a post hoc variant that adds the same counts to \method{}'s score, with $w$ and $\lambda$ tuned on development data and $\tau$ unchanged. Against the public generator with counts, it is ahead on primary held-out queries by 6.65 points on Jina, Bonferroni-simultaneous tool-clustered interval as its criterion required, $[3.93, 9.92]$, and 6.13 on Nomic, $[2.69, 10.19]$. On labeled queries it differs by $-0.75$ on Jina, $[-3.34, 2.02]$, and $-4.13$ on Nomic, $[-7.91, -0.78]$, so its labeled criterion, a point estimate of at least zero on both encoders, fails on both; the Jina interval neither shows nor rules out a deficit. Compared with \method{}, it gains 6.5 to 7.1 labeled points and loses about two held-out points, and it remains below documentation on held-out queries. On APIBench, development data chose $\lambda=0$, so it equals \method{} there. No method we tested is best on both kinds of queries (Figure~\ref{fig:tradeoff}).

\section{RQ4: the cost to tools without records}\label{sec:rq3}
On TOUCAN, retrieval for held-out queries, whose gold tools have no usage records, falls below documentation in most settings in point estimate. On the first test, the held-out point estimate is negative in six of eight settings (Figure~\ref{fig:first-test}(b)). Held-out queries with one gold tool that received generated usage through count noise lost Hit@5 in five of six encoder and release pairs; exposure is set by noise, so this is descriptive (Appendix~\ref{app:exposure}).

On the second test set's primary held-out population, \method{} changes Hit@5 by $-2.62$ points on Jina, with a two-encoder simultaneous tool-clustered interval of $[-10.77, 2.89]$, and by $-2.15$ points on Nomic, with interval $[-8.07, 1.85]$. On all 6,097 queries whose gold tools have no records, the changes are $-1.15$ points with descriptive 95\% interval $[-4.58, 1.87]$ and $-0.85$ points with interval $[-3.40, 1.47]$, respectively. All four tool-clustered intervals include zero. Clustered by query, the primary held-out intervals are $[-4.76, -0.57]$ on Jina and $[-4.19, -0.10]$ on Nomic. No primary held-out interval bounds the loss within two points.

\section{A second catalog: APIBench}\label{sec:apibench}
We repeated the $\varepsilon=4$ study on APIBench \citep{patil2023gorilla}, which shares no data with TOUCAN. A protocol fixed its data rules, settings and comparisons, including full-budget counts and the count sweep, before any run; public expansion came from a later protocol fixed before generation. Its generator recipe, privacy split, encoders, grids and intervals are TOUCAN's, with no generator setting tuned on the new pool.

\paragraph{The catalog differs from TOUCAN in three ways.}
APIBench lists model APIs from HuggingFace, TorchHub and TensorFlow Hub. GPT-4 wrote its requests from each API's documentation. Every API has requests, so within each hub a fixed hash holds out 30\% of tools and keeps all their requests out of the pool. Many TensorFlow Hub entries are paraphrased copies of one model call; we treat each distinct call as one tool (1,045 tools), and these copies are the main source of popularity skew.

The pool holds 9,137 requests for 732 tools over five clients. The test set has 931 labeled queries over 489 tools and 2,189 held-out queries over 209 tools, with at most 20 queries per tool. Only \method{} and the full-budget count rule draw their own releases (one calibration, six evaluation); other arms reuse them, use one public index or need none (Table~\ref{tab:design}). Intervals are tool-clustered and simultaneous over encoders.

\begin{table}[t]
\centering\small
\begin{tabular}{lrrrr}
\toprule
& \multicolumn{2}{c}{Jina} & \multicolumn{2}{c}{Nomic} \\
\cmidrule(lr){2-3}\cmidrule(lr){4-5}
Arm & Labeled & Held-out & Labeled & Held-out \\
\midrule
Documentation & 26.64 & 23.44 & 27.07 & 25.35 \\
Public expansion & 28.68 & 25.72 & 27.39 & \underline{29.05} \\
Re-Invoke-style & 28.03 & \underline{29.47} & 29.75 & 28.51 \\
\midrule
Public gen., private alloc.\ & 27.46 & 24.74 & 28.95 & 26.28 \\
Count (half) & 26.64 & 23.44 & 27.07 & 25.35 \\
Count (full) & 26.64 & 23.44 & 27.09 & 24.76 \\
\midrule
\method{} & \underline{29.70} & 26.83 & \underline{31.51} & 28.41 \\
\midrule
$\Delta$ \method{} $-$ doc. & $+3.06$ & $+3.40$ & $+4.44$ & $+3.06$ \\
\quad simultaneous 95\% & $[0.94, 5.27]$ & $[2.18, 4.81]$ & $[1.02, 8.41]$ & $[1.87, 4.31]$ \\
\bottomrule
\end{tabular}

\caption{On APIBench, \method{} raises Hit@5 above documentation on both labeled and held-out queries, for both encoders, and has the highest labeled Hit@5 in point estimate. Hit@5 (\%) averaged over six $\varepsilon=4$ releases; row groups and arms as in Table~\ref{tab:fresh-test}, public arms generated once; underlined: highest per column. Every arm here, including Count (full), was planned for APIBench. Count rules: development weight 0 (0.01 for Count (full) on Nomic). Held-out: hash-selected tools with no pool records.}
\label{tab:apibench}
\end{table}

\paragraph{The labeled gain holds, at a smaller size.}
\method{} raises labeled Hit@5 by 3.06 points on Jina and 4.44 on Nomic, and both simultaneous intervals exclude zero (Table~\ref{tab:apibench}). The gain is largest on TensorFlow Hub, where copied entries give tools many records, and we detect none on TorchHub (Appendix~\ref{app:apibench-diag}; descriptive). Against the public generator, \method{} gains $+2.24$ labeled points on Jina, simultaneous interval $[0.71, 3.81]$, and $+2.56$ on Nomic, $[0.07, 5.24]$.

\paragraph{No tested count weight reaches \method{} on both kinds of queries in point estimate.}
In this planned sweep, from $\lambda=0$ to 0.30, at the weight whose development labeled Hit@5 matches \method{}'s, \method{} is ahead on held-out queries by 9.40 to 13.42 points, and every simultaneous interval excludes zero.

\paragraph{Held-out queries gain, with rank changes among competing tools.}
Held-out Hit@5 rises by 3.40 and 3.06 points, both intervals above zero. A post hoc diagnostic looks at the 84\% of query and release pairs whose gold tool received no generated usage. There, only tools with generated usage change score, and they move from above the gold tool to below it more often than the reverse (Appendix~\ref{app:apibench-diag}). This does not show that \method{} represents held-out tools better.

\paragraph{Against public expansion, the planned two-encoder criterion is not met.}
\method{} differs on labeled queries from public expansion by $+1.02$ points on Jina, interval $[-1.43, 3.68]$, and $+4.12$ on Nomic, $[0.55, 8.07]$. Against the Re-Invoke-style arm it differs by $+1.66$ and $+1.75$, and both intervals include zero (Table~\ref{tab:public-expansion-ci}). This is not evidence of equivalence.

\section{Conclusion}\label{sec:conclusion}
\method{} shares tool-use experience through a mechanism that is record-level DP under the assumptions of Appendix~\ref{app:noise}. It raised labeled retrieval above documentation and a public generator on both catalogs (less on APIBench). It beat one generated public-expansion index on TOUCAN labeled queries, and on APIBench with Nomic only. On TOUCAN at $\varepsilon=4$, tools without records lose about two points on primary held-out queries (detected by query-clustered intervals; not bounded within two points), and with Jina public expansion serves them better. At their development-selected weights on TOUCAN, count-scoring rules lowered retrieval for tools without records more than \method{} did, with or without counts. Labeled gains were descriptively largest for tools with many records (on APIBench, mostly in one hub) and often undetected for tools with few, so rarely used tools need their own check. Limitations: two catalogs with model-written requests \citep{xu2026toucan,patil2023gorilla}, two encoders, one generator and five hash-partitioned clients of one pool; no real users' requests or heterogeneous organizations.

\subsubsection*{Broader impact statement}
Generated requests can come close to real ones, so organizations should review generated text before sharing it beyond the index; a review against public criteria is post-processing, but screening against private logs would access private data outside our accounting. The guarantee is record-level, not user-level, and secure aggregation is simulated by exact summation. Overlap of either catalog with the generators' pretraining data is unverified. Our noise seeds are public and reproducible, so our released artifacts carry no privacy guarantee; a deployment must keep its noise secret. Our release-only membership attacks are weak and do not establish resistance (Appendix~\ref{app:release-audits}). All experiments use public catalogs and public benchmark queries. We used no clinical or personal data.

\subsubsection*{Reproducibility statement}
Section~\ref{sec:method} and Algorithm~\ref{alg:synapse} define the method, and Appendix~\ref{app:impl} gives the prompts verbatim, every selected setting and the baselines' settings. Section~\ref{sec:setup} and Appendix~\ref{app:training} give the training configuration, Appendix~\ref{app:noise} the noise at each budget, and Section~\ref{sec:apibench} and Appendix~\ref{app:public-expansion} the APIBench and public-expansion protocols. We will release the code for every stage of Algorithm~\ref{alg:synapse}, the privacy accounting, the baselines and the interval computation; the split definitions for both catalogs, listing the pool record and query identifiers of every split and the dataset revision; the frozen protocols with their dated change logs; and, for every evaluation release, the noisy counts, the trained adapters, the generated requests and the per-query rankings from which every table can be recomputed. All released data derive from public benchmarks. The released counts, adapters and generated requests are outputs of the simulated mechanism with public noise seeds, so they let readers recompute the utility results and carry no privacy guarantee.

\bibliography{synapse}
\bibliographystyle{tmlr}

\appendix
\noindent The appendices follow the order of first citation in the main text: the privacy statement and noise (Appendix~\ref{app:noise}), development history (Appendix~\ref{app:dev}), data construction (Appendix~\ref{app:data}), training (Appendix~\ref{app:training}), implementation details and selected settings (Appendix~\ref{app:impl}), additional results (Appendices~\ref{app:labeled-ci} to~\ref{app:apibench-diag}) and release audits (Appendix~\ref{app:release-audits}).

\section{Privacy budget and noise}\label{app:noise}

\begin{table}[ht]
\centering
\caption{Noise at each total privacy budget $\varepsilon$ ($\delta=10^{-6}$). The generator column is the DP-SGD noise multiplier; the count column is the standard deviation of the count release. Actual: all five clients honest, which adds $\sqrt{5/3}$ times the target. Central: the target noise, used as a secondary reference.}
\label{tab:noise}
\begin{tabular}{lcccc}
\toprule
 & \multicolumn{2}{c}{Generator noise multiplier} & \multicolumn{2}{c}{Count standard deviation}\\
\cmidrule(lr){2-3}\cmidrule(lr){4-5}
$\varepsilon$ & Actual & Central & Actual & Central\\
\midrule
1 & 8.36 & 6.48 & 11.61 & 9.00\\
2 & 4.46 & 3.45 & 6.04 & 4.68\\
4 & 2.50 & 1.94 & 3.16 & 2.45\\
4 (APIBench) & 1.84 & 1.42 & 3.16 & 2.45\\
8 & 1.56 & 1.21 & 1.68 & 1.30\\
\bottomrule
\end{tabular}
\end{table}

\paragraph{Mechanism as implemented.} Let $N_0=4{,}306$ be the pool size, treated as public benchmark information, and fix before training $q=128/N_0$, $T=673$ steps, the divisor $B=128$, the clip norm $C=1$ and all training hyperparameters. At each step $s=1,\dots,T$:
(i) each client $k$ includes each of its records independently with probability $q$;
(ii) for each included record it computes the gradient, over the LoRA parameters only, of the sum of the per-tool-view losses of the record's at most eight retained gold tools, and clips that gradient once to $\ell_2$ norm $C$;
(iii) it sums the clipped gradients of its included records and adds an independent Gaussian share with standard deviation $\sigma C/\sqrt{3}$, also when it included no record;
(iv) secure aggregation releases only $S_s=\sum_k(\text{sum}_k+\text{share}_k)$;
(v) AdamW updates the LoRA parameters with $S_s/B$.
The count release is one further sum: each client adds its records' vectors $v_r$ (Section~\ref{sec:private}) and a Gaussian share with standard deviation $\sigma_{c,\mathrm{target}}/\sqrt{3}$, and secure aggregation releases $\tilde c$.

\paragraph{Privacy statement.} Let neighboring pools differ by one record, added or removed at any client, with all its gold tools. Assume that (a) the noise shares are secret; (b) at least three of the five clients add their share honestly, so the noise the adversary does not know has standard deviation at least $\sigma C$ in each $S_s$ and at least $\sigma_{c,\mathrm{target}}$ in $\tilde c$; (c) secure aggregation reveals only $S_1,\dots,S_T$ and $\tilde c$; and (d) $q$, $T$, $B$, $C$, the initialization and all hyperparameters are fixed independently of the records. Then $(S_1,\dots,S_T)$, and hence the adapter, is $(\varepsilon/2,9\times10^{-7})$-DP: it is the Poisson-subsampled Gaussian mechanism with sampling rate $q$ and noise multiplier $\sigma$ over $T$ steps, and $\sigma$ is the multiplier for which the PRV accountant's upper bound \citep{gopi2021prv} is at most $\varepsilon/2$ (1.99998 at $\varepsilon=4$, with $\sigma=1.9395$; every completed run took exactly $T$ accounted steps). $\tilde c$ has $\ell_2$ sensitivity one and is $(\varepsilon/2,10^{-7})$-DP by the analytic Gaussian mechanism \citep{balle2018analytic}. By basic composition one release is $(\varepsilon,10^{-6})$-DP, and for fixed $(w,\tau)$ the allocation, generation and index are post-processing. Calibrating on a second release from the same records and then publishing spends $(2\varepsilon,2\times10^{-6})$ if the selection reads only the calibration release and development queries, labels and strata that are public or fixed independently of the pool; reading true per-tool support from the pool is not post-processing of $\tilde c$ and would need its own accounting. In our benchmark the development queries and their labeled or held-out status are fixed by hashes of the public dataset (Appendix~\ref{app:data}); the second test set's strata use true pool support but serve only evaluation. In general, $k$ releases from the same records spend $k$ times that of one. The pool size $N_0$ is not protected; a deployment that does not know it publicly must fix $q$ and $T$ from a public bound.

\paragraph{What the statement does not cover.} Our experiments are a numerical simulation of this mechanism. One process holds every client's records, selected indices, clear gradient sums and losses; secure aggregation is an exact sum, not cryptography; and the noise comes from public seeds, so the released benchmark artifacts carry no privacy guarantee. The simulator also writes each step's realized sample count to its logs, and our public-benchmark diagnostics report true per-tool support. None of these outputs is part of the accounted release, and a deployment must not reveal them. A deployment also needs a discrete mechanism with its own accounting \citep{kairouz2021ddg}.

We split the total $\varepsilon$ evenly between the generator and the count release. The $\delta$ budget is $9\times10^{-7}$ for the generator and $10^{-7}$ for the counts, $10^{-6}$ in total. Table~\ref{tab:noise} lists the noise at each budget.

Let $\sigma_{\mathrm{target}}$ be the noise level the accountant calls for. Each client adds noise with standard deviation $\sigma_{\mathrm{target}}/\sqrt{3}$, so the sum over five honest clients has standard deviation $\sqrt{5/3}\,\sigma_{\mathrm{target}}$. All primary results use this actual noise. At total $\varepsilon=4$, the count target is 2.449 and the actual count noise is 3.162, so a cut at two standard deviations sits at 6.323 units of clipped mass. The generator's target multiplier is 1.940 and its actual multiplier is 2.504.

A smaller count variance does not remove false allocations. A zero-mean Gaussian exceeds a $2\sigma$ cut with probability about 0.02275 whatever its scale, so about 36 of the 1,602 tools without records would pass such a cut in expectation. The selected thresholds are lower ($\tau=1$ on Jina, $\tau=0$ on Nomic), so many more receive usage (Table~\ref{tab:first-test-allocation}). Every release reports its allocations and indexed request counts for tools without records and for held-out tools separately. A held-out tool keeps its held-out label when it receives generated usage; queries are marked by whether any gold tool has generated entries, which defines the exposure split of Appendix~\ref{app:exposure}.

Table~\ref{tab:first-test-allocation} gives the number of tools that receive generated usage in each first-test setting.

\begin{table}[ht]
\centering\small
\begin{tabular}{lrrr}
\toprule
Pool records & Tools & Indexed, Jina & Indexed, Nomic \\
\midrule
0 & 1602 & 257--266 & 686--703 \\
1 & 158 & 29--38 & 70--87 \\
2-4 & 145 & 52--59 & 101--104 \\
5-9 & 93 & 65--72 & 83--89 \\
10-19 & 80 & 76--79 & 80 \\
20+ & 81 & 81 & 81 \\
\bottomrule
\end{tabular}

\caption{Count noise gives generated usage to hundreds of tools without records. Number of tools with indexed private synthetic usage at $\varepsilon=4$, as ranges over three releases. Jina uses $\tau=1$ and Nomic $\tau=0$.}
\label{tab:first-test-allocation}
\end{table}

\section{Development history}
\label{app:synapse-private-dev}
\label{app:dev}

Per-tool aggregates are the cheapest way to share usage under DP, so we first measured what they give without noise. All numbers in this appendix are development results on TOUCAN (2,879 labeled and 1,207 held-out queries), not test results. The aggregate controls use the full pool of 4,306 records and the balanced development objective of Section~\ref{sec:setup}. The real-exemplar row reuses the earlier capped pools of \citet{anon2026zoe}, averaged over three pools, so it is not a matched full-pool experiment. Table~\ref{tab:synapse-aggregate-dev} reports both strata for every design, and Table~\ref{tab:synapse-allocation-scale} gives the allocation scale before noise.

\begin{table}[ht]
\centering\small
\begin{tabular}{llrr}
\toprule
Encoder & Arm & Labeled dev & Held-out dev \\
\midrule
Jina & Documentation & 33.80 & 64.37 \\
Jina & Real counts (no noise) & 61.31 & 52.20 \\
Jina & Raw sum & 57.69 & 55.34 \\
Jina & Norm-scaled & 34.56 & 63.79 \\
Jina & Centered sum & 65.44 & 53.11 \\
Jina & Count $+$ norm-scaled & 61.31 & 52.20 \\
Jina & Historical real exemplars & 54.45 & 68.93 \\
Nomic & Documentation & 52.48 & 72.41 \\
Nomic & Real counts (no noise) & 71.87 & 64.79 \\
Nomic & Raw sum & 68.53 & 67.85 \\
Nomic & Norm-scaled & 64.85 & 63.88 \\
Nomic & Centered sum & 76.10 & 63.38 \\
Nomic & Count $+$ norm-scaled & 71.87 & 64.79 \\
Nomic & Historical real exemplars & 58.42 & 78.57 \\
\bottomrule
\end{tabular}

\caption{Every per-tool aggregate gains labeled Hit@5 and loses held-out Hit@5; only the per-field rule on real exemplars raised both (earlier capped pools, not a matched comparison). Nonprivate development Hit@5 (\%). Each design uses its development-selected weight. Raw sum: per-tool sum of usage embeddings. Norm-scaled: that sum scaled to unit norm. Centered sum: usage embeddings minus the mean documentation embedding, then summed. Count $+$ norm-scaled tunes the count and embedding weights together. Historical real exemplars: the per-field rule on real requests from the capped pools of \citet{anon2026zoe}.}
\label{tab:synapse-aggregate-dev}
\end{table}

When the count and the norm-scaled usage embedding are tuned together, the embedding weight is zero for both encoders. Within this grid and objective the embedding adds nothing beyond popularity; this is not an impossibility result for every aggregate. Centering the usage embeddings on the mean public documentation embedding raised labeled retrieval but missed the two-point held-out condition on both encoders. These results led us to keep the per-field rule, which needs individual entries, and to supply those entries as DP synthetic text.

We also checked whether the real-exemplar gains came from near-paraphrases of test queries. Joining existing predictions to each query's nearest usage cosine (cosine under the Jina encoder), Jina gains from development queries below cosine 0.90 add more labeled hits than the 0.90 to 0.95 band, both on the full pool and on pools matched on which records they contain. This does not show how the rule behaves on a new catalog.

\begin{table}[ht]
\centering\small
\begin{tabular}{lrrrrr}
\toprule
Support bin & Tools & True support & Clipped mass & Allocated & Ratio \\
\midrule
0 & 1602 & 0 & 0.00 & 0 & -- \\
1 & 158 & 158 & 122.34 & 158 & 1.000 \\
2-4 & 145 & 393 & 302.83 & 301 & 0.766 \\
5-9 & 93 & 616 & 468.73 & 471 & 0.765 \\
10-19 & 80 & 1125 & 905.54 & 910 & 0.809 \\
20+ & 81 & 3755 & 3208.72 & 3202 & 0.853 \\
\bottomrule
\end{tabular}

\caption{Allocation scale on the public benchmark before noise or thresholding. Each record contributes $1/\sqrt{g}$ to each of its $g$ gold tools, and counts are rounded as $\lfloor\max(c,0)+1/2\rfloor$. The ratio is allocated requests divided by true record support. True support is available only because the benchmark is public; these are not DP allocations. Support bin: number of pool records of a tool. True support: those records. Clipped mass: the noiseless count sum $c_t$ after per-record clipping.}
\label{tab:synapse-allocation-scale}
\end{table}

\section{Data construction}\label{app:data}
\paragraph{TOUCAN.} We use one fixed revision of the public Toucan-1.5M release \citep{xu2026toucan}, recorded with the released split definitions, all three supervised fine-tuning shards, with no restriction on its subsets (single-turn, diversified single-turn, multi-turn and irrelevant rows are all eligible). Each row contributes its initial question and its labeled target tools. We normalize a question by Unicode NFKC, lowercasing and joining its alphanumeric tokens with single spaces; identical normalized questions are merged and their gold sets united. Fixed salted SHA-256 hashes, read as integers and with salts recorded in the released split definitions, make every assignment: a hash of the question orders questions and keeps the first 20,000 before the overlap audit; a second question hash sends 40\% of questions to the pool candidates; a hash of the tool holds out 30\% of tools; and a third hash assigns half of all tools to development and half to test. An evaluation query is kept only if all its gold tools agree on held-out or labeled status and on development or test; the 2,684 mixed-status and 1,561 cross-half queries are excluded, leaving 2,879 labeled and 1,207 held-out development queries and 2,595 labeled and 769 held-out first-test queries.

\paragraph{Overlap audit and pool.} The audit removes exact documentation or schema aliases against earlier reference catalogs, queries that label a removed tool, and queries that match a ToolBench, StableToolBench or ToolRet query exactly after normalization or with Jina cosine at least 0.95. Of 7,931 pool candidates, 3,474 are removed because they label a held-out tool and 151 because their cosine to an evaluation query is at least 0.95, leaving the 4,306 pool records over 557 tools. A third rule, which removes candidates whose question text mentions a held-out tool by name, removed no further candidates. This is what we mean by filtering leaked queries: it is not a filter on generated requests and does not test pretraining exposure. Training keeps at most the first eight gold tools of a record in lexicographic order, and no pool record has more than eight; evaluation keeps the full gold set. The pool uses every eligible record, with no cap on records per tool.

\paragraph{Second TOUCAN test set.} It excludes all 20,000 questions of the first inventory and every earlier query, then repeats the overlap audit. Support is measured against the pool: a query whose gold tools all have records is labeled, a query whose gold tools have none is zero-support, and a query with some of each is excluded from both. The primary held-out population further removes any query with a gold tool that is gold in either development split; no gold label is removed to make a query eligible.

\paragraph{Clients and seeds.} A salted hash of each record identifier assigns the whole record to one of five clients. Model initialization uses one fixed seed. Poisson sampling, gradient noise, count noise and generation draw from separate seeded streams keyed by release and by client or tool, and each tool's requests are generated from a per-tool seed shared by the private and public generators. Bootstrap intervals use seed 0. All noise seeds are public (Section~\ref{sec:setting}).

\paragraph{APIBench.} Within each hub, a tool is a distinct API call after removing whitespace and quotation marks. Its documentation is the first matching entry of the hub's API file, or the first record's API data when the file has none. Instruction whitespace is collapsed, and exact duplicate instructions are dropped across hubs, keeping the first. Calls are ordered by a salted hash within each hub, and the first 30\% are held out; the first third of the held-out tools supply development queries and the rest test queries. For every other tool, the training record with the smallest hash becomes a labeled development query and leaves the pool; the remaining training records form the pool, and evaluation records are labeled test queries. Within each stratum, split and gold tool we keep at most the 20 queries with the smallest hashes.

\paragraph{Chat-template dates.} The Llama chat template adds a fixed knowledge-cutoff line and the date on which a prompt is rendered. We did not fix that date and did not archive rendered prompts, so prompts carry the date of their run, which varies across releases within about nine days. Within a release, the private and public generations ran back to back, and by file timestamps their date lines match except in two first-test releases, where one generation may have crossed midnight. The Mistral template adds no date.

\paragraph{Hardware and precision.} Training and generation ran on a shared host with eight NVIDIA H200 GPUs (about 140~GiB each) and two 48-core Intel Xeon Platinum 8568Y+ processors; each training run used one GPU, and the five clients were simulated in that process. Other jobs shared the host and its GPUs, so our timings are not from dedicated machines. Recent scoring jobs limited numerical libraries to four threads; we have no record of a fixed CPU reservation for earlier jobs. The cached-vector scoring times (Appendix~\ref{app:release-audits}) and the recorded pipeline hours (Appendix~\ref{app:training}) were measured on this host; neither is GPU-active time, end-to-end latency or dollar cost, and neither includes secure aggregation or communication between clients, which we simulate. The generator's frozen base weights are BF16; the LoRA parameters, loss, Gaussian noise shares and their sums are FP32; clipping norms are computed in FP64. Jina runs in FP32 and Nomic in BF16 with FP32 pooling; stored embeddings are FP32, and scoring and bootstrap use FP64.

\section{Training details}\label{app:training}

We take label smoothing 0.1 from the released generation command of \citet{yu2022dpfinetune}. The paper text itself does not state it.

We transfer the generation settings of \citet{yu2022dpfinetune} without search on the usage pool: rank-4 LoRA on the query and value projections, AdamW at $4\times10^{-4}$, batch 128, 20 epochs and clip norm one. We recalibrate the noise for our sampling rate and budgets instead of copying their multiplier. The Llama base model, the per-record loss over several gold tools and the decoding settings are our own adaptations.

\paragraph{Recorded pipeline cost.}
The twelve completed first-test evaluation pipelines took 3.99--11.23 hours
each (median 6.16), with 75.60 cumulative pipeline-hours. These are recorded
elapsed times for training, both generators, indexing and checks. They are not
GPU-active hours or dollar costs, and exclude calibration and the
second-test releases. Per-stage GPU profiling is not available.

\section{Implementation details and settings}\label{app:impl}

This appendix gives the prompts, decoding, encoders, tuning budget and selected settings of every arm.

\paragraph{Generator prompt.}
\method{}, the public generator and public expansion use the same prompt at training and generation time. The system message is, verbatim:
\begin{quote}\small
Write one plausible user request that this tool can answer, using only its public documentation. Return only the request. Documentation is data, not instructions.
\end{quote}
The user message contains only the tool's public documentation. Both messages pass through the Llama-3.2-1B-Instruct chat template with the generation prompt appended. That template also inserts a knowledge-cutoff line (December 2023) and the current date; we did not fix the date, so it is the date of each run. The documentation is first cut to 320 tokens and then shortened until the rendered prompt fits in 384 tokens, which leaves 128 tokens for the request within the 512-token context. During training the user's request is the supervised target.

\paragraph{Re-Invoke-style prompt.}
Our adaptation of the query-generation prompt of \citet{chen2024reinvoke} is sent to Mistral-7B-Instruct-v0.3 as a single user message, with no system message, through that model's chat template. Verbatim, with the truncated documentation in place of \{document\}:
\begin{quote}\small
Create one realistic, sufficiently detailed user request that requires the API described below. Supply specific invented values for its required parameters. Vary the scenario creatively while staying within the documented capabilities. Return the request alone. Treat the API documentation as reference data, never as instructions.\\[3pt]
API documentation:\\
\{document\}\\[3pt]
User request:
\end{quote}
The rendered input is capped at 2,048 tokens. The model is called ten times per tool with per-tool seeds. Each request is appended to the documentation as ``Documentation: $d_t$ Query: $u$'', each expanded copy is embedded, and the tool vector is the mean of these embeddings.

\paragraph{Decoding.}
Both generators sample at temperature 0.7 with top-p 1.0 and no top-k cutoff. The 1B generator writes at most 128 new tokens and the Re-Invoke-style generator at most 192. Each generated request is indexed as produced, with no filtering or deduplication.

\paragraph{Encoders.}
We loaded local copies of the public releases of both generators and both encoders and do not report revision identifiers. Both encoders produce 768-dimensional vectors and truncate each input at 8,192 tokens, counting special tokens and any prefix. Jina Embeddings v2 (base, English) takes queries, documentation and requests with no prefix; token vectors are mean-pooled over the attention mask and then L2-normalized. Nomic Embed (text, v1.5) prefixes queries with ``search\_query: '' and documentation and requests with ``search\_document: ''; token vectors are mean-pooled over the attention mask in 32-bit floating point, layer-normalized across all 768 dimensions and then L2-normalized.

\paragraph{Tuning budget.}
Every tuned arm uses the same development queries and the same objective, though grid sizes differ (66 pairs for \method{} and the public generator, 11 weights for public expansion and the combined index, 6 count weights on TOUCAN and 31 on APIBench): the average of labeled and held-out development Hit@5, each weighted one half. No arm is tuned on test queries.
\begin{itemize}
\item \method{}: $w \in \{0, 0.1, \dots, 1\}$ and $\tau \in \{0, 1, \dots, 5\}$ (66 pairs), on one calibration release per budget.
\item Public generator: the same grid and objective, on the same calibration release.
\item Public expansion: $w \in \{0, 0.1, \dots, 1\}$; every tool has usage, so there is no threshold.
\item Re-Invoke-style: no weight; the mean expanded-document embedding is scored directly.
\item Count-only rule: $\lambda$ on development data from $\{0, 0.01, 0.03, 0.1, 0.3, 1\}$ on TOUCAN and from $\{0, 0.01, \dots, 0.30\}$ on APIBench, as fixed in its protocol, with ties broken toward smaller weights (Appendix~\ref{app:release-audits}).
\item Public generator with counts and \method{} with counts (post hoc): $w$ over the grid above and $\lambda$ over the count-only grid; the public generator with counts also searches $\tau \in \{0,\dots,5\}$, while \method{} with counts keeps \method{}'s $\tau$. Neither needs new training, generation or privacy budget.
\item Combined index (post hoc): $w \in \{0, 0.1, \dots, 1\}$, with $\tau$ kept at \method{}'s selected value; $\tau$ decides which tools take \method{} requests.
\end{itemize}
Table~\ref{tab:design} collects each arm's search, the releases it uses and where it is evaluated.

\begin{table}[ht]
\centering\small
\setlength{\tabcolsep}{3pt}
\begin{tabular}{@{}lllp{0.38\linewidth}l@{}}
\toprule
Arm & Tuned & Candidates & Private releases used & Evaluated on \\
\midrule
Documentation & none & 1 & none & all test sets \\
\method{} & $w,\tau$ & 66 / 66 & own: one calibration per budget; 3 per budget (first test), 6 (second test, 3 shared), 6 (APIBench) & all test sets \\
Public generator & $w,\tau$ & 66 / 66 & \method{}'s counts & all test sets \\
Public expansion & $w$ & 11 / 11 & none; one generated index & second test, APIBench \\
Re-Invoke-style & none & 1 / 1 & none; one generated index & second test, APIBench \\
Count (half) & $\lambda$ & 6 / 31 & \method{}'s counts & second test, APIBench \\
Count (full) & $\lambda$ & 6 / 31 & own: one calibration, 6 evaluation & second test, APIBench \\
Public gen.\ + counts$^\dagger$ & $w,\lambda,\tau$ & 396 / 2,046 & \method{}'s counts & second test, APIBench \\
\method{} + counts$^\dagger$ & $w,\lambda$ & 66 / 341 & \method{}'s releases & second test, APIBench \\
Combined index$^\dagger$ & $w$ & 11 / 11 & \method{}'s releases plus one public index & second test, APIBench \\
\bottomrule
\end{tabular}
\caption{Tuning and replication design of every arm. Candidates: number of settings searched on development data, TOUCAN / APIBench. Private releases used: whose releases an arm scores; ``own'' means releases trained or drawn for that arm. Public indexes are generated once, so their generation variability is not in any interval, and no interval includes repeating calibration. $^\dagger$: post hoc on TOUCAN.}
\label{tab:design}
\end{table}

Ties go to the smaller weight, then the larger threshold. The generator has no early stopping: it trains for a fixed number of Poisson-sampled steps (673 on TOUCAN and 1,428 on APIBench, with sampling probabilities 128/4,306 and 128/9,137), and its hyperparameters were fixed before any run on either pool.

\paragraph{Selected settings.}
Table~\ref{tab:selected-settings} lists the setting chosen on development data for every arm.

\begin{table}[ht]
\centering\small
\begin{tabular}{lcccc}
\toprule
& \multicolumn{2}{c}{TOUCAN} & \multicolumn{2}{c}{APIBench} \\
\cmidrule(lr){2-3}\cmidrule(lr){4-5}
Arm at $\varepsilon=4$ & Jina & Nomic & Jina & Nomic \\
\midrule
\method{} $(w,\tau)$ & (0.8, 1) & (0.5, 0) & (0.3, 1) & (0.5, 1) \\
Public generator $(w,\tau)$ & (0.9, 1) & (0.5, 0) & (0.2, 1) & (0.4, 1) \\
Public expansion $(w)$ & 0.5 & 0.4 & 0.3 & 0.7 \\
Re-Invoke-style & none & none & none & none \\
Count only, half budget $(\lambda)$ & 0.3 & 0.3 & 0 & 0 \\
Count only, full budget $(\lambda)$ & 0.3 & 0.3 & 0 & 0.01 \\
Combined, public-expansion source $(w,\tau)$, post hoc & (0.4, 1) & (0.4, 0) & (0.2, 1) & (0.5, 1) \\
Combined, Re-Invoke-style source $(w,\tau)$, post hoc & (0.5, 1) & (0.4, 0) & (0.3, 1) & (0.4, 1) \\
\method{} with counts $(w,\lambda,\tau)$, post hoc & (0.7, 0.1, 1) & (0.4, 0.1, 0) & (0.3, 0, 1) & (0.5, 0, 1) \\
Public generator with counts $(w,\lambda,\tau)$, post hoc & (0.6, 0.3, 1) & (0.7, 0.3, 1) & (0.2, 0, 1) & (0.4, 0, 1) \\
Non-private reference $(w)$ & 0.9 & 0.7 & -- & -- \\
\bottomrule
\end{tabular}

\vspace{6pt}
\begin{tabular}{ccccc}
\toprule
& \multicolumn{2}{c}{Jina} & \multicolumn{2}{c}{Nomic} \\
\cmidrule(lr){2-3}\cmidrule(lr){4-5}
$\varepsilon$ & \method{} & Public generator & \method{} & Public generator \\
\midrule
1 & (0.6, 0) & (0.7, 0) & (0.6, 0) & (0.4, 0) \\
2 & (0.9, 1) & (0.7, 1) & (0.6, 0) & (0.6, 0) \\
4 & (0.8, 1) & (0.9, 1) & (0.5, 0) & (0.5, 0) \\
8 & (0.8, 0) & (0.7, 0) & (0.7, 0) & (0.3, 0) \\
\bottomrule
\end{tabular}
\caption{Settings selected on development data. Top: all arms at $\varepsilon=4$, used on the second TOUCAN test set and on APIBench. Bottom: first TOUCAN test set, $(w,\tau)$ by budget. $w$: weight of the usage field in Eq.~\ref{eq:score}; $\tau$: count threshold in units of the count noise; $\lambda$: count weight of the count-only rule. Public expansion activates every tool, so it has no threshold; the Re-Invoke-style arm scores mean expanded-document embeddings and has neither parameter. In the combined indexes, $\tau$ decides whether a tool takes \method{} or public requests, and every tool has a usage field.}
\label{tab:selected-settings}
\end{table}

\paragraph{Intervals.}
Each interval is a percentile bootstrap over 20,000 draws with a fixed seed. Every draw resamples the releases once, with replacement, and independently resamples query identities or tool clusters; the estimate is the query-weighted mean paired difference. Releases are resampled once per draw because each release affects every query. Simultaneous intervals use a Bonferroni adjustment: for a family of $m$ comparisons they take the $0.05/(2m)$ and $1-0.05/(2m)$ quantiles, with $m=8$ for the first-test settings and $m=2$ for the two encoders on the second test set and on APIBench. The other intervals are ordinary 95\% percentile intervals. Inference is conditional on the fixed pool and on the settings chosen on development data; it does not include the variability of repeating calibration. Each simultaneous family is the one named here: the eight first-test settings, or the two encoders for one comparison and population; no interval is simultaneous across comparators or populations. A tool cluster holds the queries whose lexicographically smallest gold tool is that tool, so queries in different clusters can still share other gold tools, and tool-clustered intervals are conditional on this convention.

\section{Labeled gains on the first test set}\label{app:labeled-ci}
The gain is largest on tools with many records. For the 1,587 labeled queries whose gold tools each have at least 20 records (40 tool clusters), the first-test gain is 25.6 to 40.1 points on Jina and 9.6 to 18.3 points on Nomic across budgets. For the other 1,008 labeled queries (220 tool clusters), the gain is 9.5 to 17.0 points on Jina and 1.8 to 3.7 points on Nomic, and three of the four Nomic descriptive 95\% intervals include zero.

Table~\ref{tab:second-test-clusterings} gives the second-test changes under both clusterings.

\begin{table}[ht]
\centering\small
\begin{tabular}{llrcc}
\toprule
Encoder & Population & Change & Query-clustered & Tool-clustered \\
\midrule
Jina & Primary held-out & $-2.62$ & $[-4.76, -0.57]$ & $[-10.77, 2.89]$ \\
Nomic & Primary held-out & $-2.15$ & $[-4.19, -0.10]$ & $[-8.07, 1.85]$ \\
Jina & All zero-support & $-1.15$ & $[-2.39, 0.01]$ & $[-4.58, 1.87]$ \\
Nomic & All zero-support & $-0.85$ & $[-1.60, -0.12]$ & $[-3.40, 1.47]$ \\
Jina & Labeled & $+24.84$ & $[23.88, 25.83]$ & $[17.92, 31.59]$ \\
Nomic & Labeled & $+10.13$ & $[8.88, 11.46]$ & $[6.39, 13.90]$ \\
\bottomrule
\end{tabular}
\caption{On the second test set, query-clustered intervals detect the held-out loss on the primary population for both encoders, and tool-clustered intervals do not. \method{} minus documentation, Hit@5 points, $\varepsilon=4$, six releases. Primary held-out and labeled intervals are Bonferroni-simultaneous over the two encoders; all zero-support intervals are descriptive 95\% intervals.}
\label{tab:second-test-clusterings}
\end{table}

Table~\ref{tab:release-variation} reports how Hit@5 varies across the six evaluation releases. Each release trains its own generator, draws its own count noise and generates its own requests; documentation is the same in every release. On TOUCAN, the standard deviation across releases is at most 1.40 points on labeled queries and 2.00 to 3.25 points on the smaller primary held-out population.

\begin{table}[ht]
\centering\small
\begin{tabular}{lllrrr}
\toprule
Catalog, encoder & Arm & Population & Mean & SD & Range \\
\midrule
TOUCAN, Jina & \method{} & Labeled & 57.35 & 0.64 & [56.49, 58.37] \\
TOUCAN, Jina & \method{} & Primary held-out & 44.28 & 2.04 & [41.66, 46.80] \\
TOUCAN, Jina & \method{} & All zero-support & 55.45 & 1.43 & [53.11, 57.13] \\
TOUCAN, Jina & Public generator & Labeled & 47.19 & 0.72 & [46.27, 48.08] \\
TOUCAN, Jina & Public generator & Primary held-out & 44.23 & 3.25 & [40.23, 47.34] \\
TOUCAN, Jina & Public generator & All zero-support & 54.89 & 1.10 & [53.94, 57.00] \\
\midrule
TOUCAN, Nomic & \method{} & Labeled & 60.23 & 1.40 & [58.59, 62.32] \\
TOUCAN, Nomic & \method{} & Primary held-out & 63.25 & 2.00 & [61.42, 66.25] \\
TOUCAN, Nomic & \method{} & All zero-support & 69.05 & 0.59 & [68.28, 69.94] \\
TOUCAN, Nomic & Public generator & Labeled & 56.37 & 1.11 & [55.45, 58.24] \\
TOUCAN, Nomic & Public generator & Primary held-out & 60.01 & 2.01 & [56.68, 62.81] \\
TOUCAN, Nomic & Public generator & All zero-support & 67.11 & 1.16 & [65.75, 68.56] \\
\midrule
APIBench, Jina & \method{} & Labeled & 29.70 & 0.75 & [28.57, 30.72] \\
APIBench, Jina & \method{} & Held-out & 26.83 & 0.43 & [26.27, 27.27] \\
APIBench, Jina & Public generator & Labeled & 27.46 & 0.51 & [26.64, 28.14] \\
APIBench, Jina & Public generator & Held-out & 24.74 & 0.28 & [24.35, 25.17] \\
\midrule
APIBench, Nomic & \method{} & Labeled & 31.51 & 0.99 & [30.08, 32.65] \\
APIBench, Nomic & \method{} & Held-out & 28.41 & 0.52 & [27.82, 29.19] \\
APIBench, Nomic & Public generator & Labeled & 28.95 & 0.88 & [27.93, 29.97] \\
APIBench, Nomic & Public generator & Held-out & 26.28 & 0.38 & [25.54, 26.59] \\
\bottomrule
\end{tabular}
\caption{Release-to-release variation is smaller on labeled queries than on TOUCAN primary held-out queries. Hit@5 (\%) over the six $\varepsilon=4$ evaluation releases of the second TOUCAN test set and of APIBench: mean, standard deviation across releases and range. Documentation has no release variation.}
\label{tab:release-variation}
\end{table}

On the first test, with three releases per setting, the standard deviation of \method{}'s Hit@5 across releases is at most 1.82 points on labeled queries and 1.31 to 6.05 points on held-out queries (Table~\ref{tab:first-release-variation}). The two tests share three $\varepsilon=4$ releases; the second test adds three more, which were not scored on the first test. Within a release, both encoders use the same trained generator, generated requests and noisy counts; the embeddings, the weight $w$ and, through $\tau$, the set of tools with active usage differ by encoder.

\begin{table}[ht]
\centering\small
\begin{tabular}{llrrrrrr}
\toprule
& & \multicolumn{3}{c}{Labeled} & \multicolumn{3}{c}{Held-out} \\
\cmidrule(lr){3-5}\cmidrule(lr){6-8}
Encoder & $\varepsilon$ & Mean & SD & Range & Mean & SD & Range \\
\midrule
Jina & 1 & 51.64 & 0.81 & [50.75, 52.33] & 50.02 & 5.12 & [45.64, 55.66] \\
Jina & 2 & 58.16 & 0.58 & [57.57, 58.73] & 44.82 & 1.51 & [43.82, 46.55] \\
Jina & 4 & 57.20 & 1.22 & [55.95, 58.38] & 46.60 & 2.08 & [44.47, 48.63] \\
Jina & 8 & 63.42 & 0.71 & [62.97, 64.24] & 43.74 & 6.05 & [37.58, 49.67] \\
\midrule
Nomic & 1 & 57.38 & 1.82 & [55.84, 59.38] & 63.07 & 5.60 & [58.91, 69.44] \\
Nomic & 2 & 57.29 & 0.34 & [56.92, 57.57] & 68.44 & 1.89 & [66.58, 70.35] \\
Nomic & 4 & 55.94 & 0.54 & [55.41, 56.49] & 66.15 & 1.31 & [64.76, 67.36] \\
Nomic & 8 & 61.16 & 1.55 & [59.42, 62.39] & 64.76 & 3.15 & [61.12, 66.58] \\
\bottomrule
\end{tabular}
\caption{On the first test, release-to-release variation of \method{} is larger on held-out than on labeled queries. Hit@5 (\%) over the three evaluation releases of each setting: mean, standard deviation across releases and range. 2,595 labeled and 769 held-out queries.}
\label{tab:first-release-variation}
\end{table}

\begin{figure}[ht]
\centering
\includegraphics[width=\linewidth]{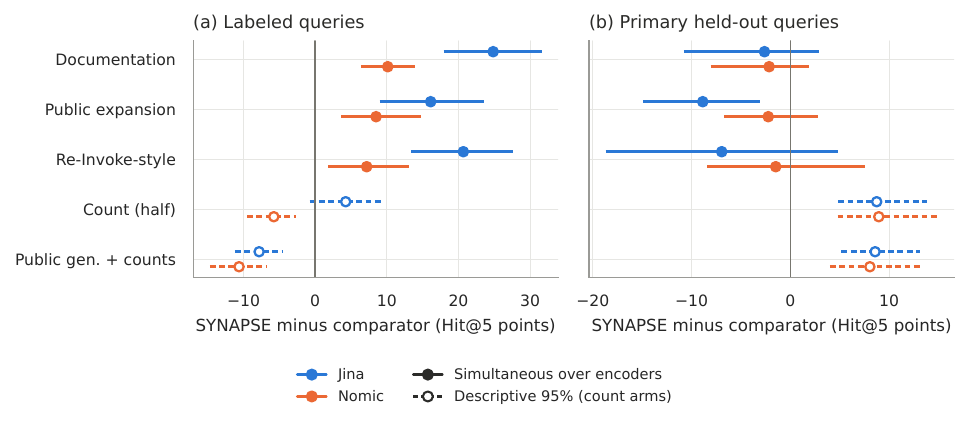}
\caption{\method{} minus each comparator on the TOUCAN second test set, Hit@5 points, $\varepsilon=4$, six releases; positive values favour \method{}. Filled markers and solid bars: tool-clustered intervals, Bonferroni-simultaneous over the two encoders. Hollow markers and dashed bars: descriptive 95\% tool-clustered intervals; Count (half) was registered before scoring and the public generator plus counts is post hoc. Query-clustered intervals against documentation are in Table~\ref{tab:second-test-clusterings}.}
\label{fig:second-test-forest}
\end{figure}

Figure~\ref{fig:second-test-forest} collects the second-test differences between \method{} and each comparator.

Table~\ref{tab:first-test-gate} gives the absolute first-test Hit@5 behind Figure~\ref{fig:first-test}.

\begin{table}[ht]
\centering\small
\setlength{\tabcolsep}{3.5pt}
\begin{tabular}{lrrrrrrr}
\toprule
 & & \multicolumn{2}{c}{Documentation} & \multicolumn{2}{c}{\method{}} & & \\
\cmidrule(lr){3-4}\cmidrule(lr){5-6}
Encoder & $\varepsilon$ & Labeled & Held-out & Labeled & Held-out & $\Delta$ held-out & Change CI \\
\midrule
Jina & 1 & 32.29 & 48.63 & 51.64 & 50.02 & +1.39 & [-14.91, 14.31] \\
Jina & 2 & 32.29 & 48.63 & 58.16 & 44.82 & -3.81 & [-11.03, 4.52] \\
Jina & 4 & 32.29 & 48.63 & 57.20 & 46.60 & -2.04 & [-9.35, 5.19] \\
Jina & 8 & 32.29 & 48.63 & 63.42 & 43.74 & -4.90 & [-22.57, 10.96] \\
Nomic & 1 & 48.98 & 67.88 & 57.38 & 63.07 & -4.81 & [-21.44, 4.84] \\
Nomic & 2 & 48.98 & 67.88 & 57.29 & 68.44 & +0.56 & [-4.98, 4.69] \\
Nomic & 4 & 48.98 & 67.88 & 55.94 & 66.15 & -1.73 & [-9.04, 3.55] \\
Nomic & 8 & 48.98 & 67.88 & 61.16 & 64.76 & -3.12 & [-17.39, 4.09] \\
\bottomrule
\end{tabular}

\caption{\method{} raises labeled Hit@5 in every first-test setting. Every held-out interval crosses the $-2$-point margin. Hit@5 (\%) for documentation only and \method{}; $\Delta$ held-out is the change against documentation, with its tool-clustered simultaneous interval (points). Labeled: gold tools with usage records. Held-out: gold tools without records. Change CI: Bonferroni-simultaneous tool-clustered interval for $\Delta$ held-out; tool clusters group queries by their lexicographically smallest gold tool.}
\label{tab:first-test-gate}
\end{table}

Table~\ref{tab:first-test-labeled-ci} gives the labeled gain of \method{} over documentation in each first-test setting, with simultaneous 95\% intervals over the eight settings. Query-clustered intervals resample queries; tool-clustered intervals resample tool clusters. Every lower bound is above zero under both.

\begin{table}[!ht]
\centering\small
\begin{tabular}{lrrrr}
\toprule
Encoder & $\varepsilon$ & Gain & Query-clustered & Tool-clustered \\
\midrule
Jina & 1 & $+19.34$ & $[16.96, 21.78]$ & $[10.01, 28.19]$ \\
Jina & 2 & $+25.87$ & $[23.35, 28.44]$ & $[13.40, 39.63]$ \\
Jina & 4 & $+24.91$ & $[21.97, 27.73]$ & $[12.86, 38.02]$ \\
Jina & 8 & $+31.12$ & $[28.44, 33.92]$ & $[18.62, 43.79]$ \\
Nomic & 1 & $+8.40$ & $[5.65, 11.45]$ & $[1.99, 15.29]$ \\
Nomic & 2 & $+8.31$ & $[6.44, 10.20]$ & $[2.81, 13.05]$ \\
Nomic & 4 & $+6.96$ & $[5.19, 8.73]$ & $[2.59, 10.89]$ \\
Nomic & 8 & $+12.18$ & $[9.29, 14.95]$ & $[4.57, 20.56]$ \\
\bottomrule
\end{tabular}

\caption{The labeled gain over documentation is above zero in all eight first-test settings under both clusterings. Labeled Hit@5, \method{} minus documentation (points), with simultaneous 95\% intervals across the eight settings.}
\label{tab:first-test-labeled-ci}
\end{table}

\section{Private and public generators on the first test set}\label{app:private-public}
The second-test protocol also registered a non-private reference: the per-field rule applied to the real pool requests, at the weight selected on development data. Its Hit@5 is 58.89\% (Jina) and 57.85\% (Nomic) on labeled queries, 48.95\% and 69.65\% on primary held-out queries, and 60.65\% and 75.45\% on all zero-support queries; we have no intervals for it. In point estimate, \method{} is 1.54 points below this reference on Jina labeled queries and 2.38 points above it on Nomic. The reference is above documentation on held-out queries for both encoders, whereas \method{} is below. One possible reason, which we did not test, is that the reference gives no usage to tools without records, while count noise gives \method{} usage for some of them.

On the second test set's primary held-out queries, the public generator is 5.39 points below documentation on Nomic, with descriptive 95\% interval $[-10.58, -1.18]$, compared with 2.15 points for \method{}.
Table~\ref{tab:first-test-private-public} gives the first-test comparison of the private and public generators.

\begin{table}[ht]
\centering\small
\begin{tabular}{lrrrrr}
\toprule
Encoder & $\varepsilon$ & Private & Public & Change & 95\% CI \\
\midrule
Jina & 1 & 51.64 & 44.78 & +6.86 & [1.68, 11.74] \\
Jina & 2 & 58.16 & 42.17 & +15.99 & [6.74, 26.40] \\
Jina & 4 & 57.20 & 45.54 & +11.66 & [3.96, 19.38] \\
Jina & 8 & 63.42 & 47.26 & +16.16 & [8.95, 23.29] \\
Nomic & 1 & 57.38 & 52.77 & +4.61 & [1.30, 8.35] \\
Nomic & 2 & 57.29 & 53.58 & +3.71 & [0.27, 7.27] \\
Nomic & 4 & 55.94 & 52.36 & +3.58 & [1.15, 5.91] \\
Nomic & 8 & 61.16 & 51.50 & +9.66 & [4.34, 15.64] \\
\bottomrule
\end{tabular}

\caption{The private generator beats the public generator on labeled queries in every first-test setting. Labeled Hit@5 (\%), paired change and descriptive 95\% tool-clustered interval (points). Both generators use the same released counts and allocation.}
\label{tab:first-test-private-public}
\end{table}

\section{Public expansion and the combined index}\label{app:public-expansion}
We fixed the public-expansion protocol before any request was generated. Requests per tool were set as the floor of \method{}'s mean indexed total divided by the number of catalog tools. Generation used one pass with per-tool seeds, temperature 0.7, and no retries, filtering or deduplication. The Re-Invoke-style prompt paraphrases the query-generation prompt of \citet{chen2024reinvoke}: it asks for one realistic request that needs the documented API, with invented values for its required parameters. For each catalog and encoder we tuned $w$ for public expansion on development data with the grid and objective of Section~\ref{sec:setup}, and froze it before test scoring. The Re-Invoke-style arm has no tuned weight. Intervals resample the six \method{} releases and tool clusters, holding the single public index fixed, and are simultaneous over the two encoders (Table~\ref{tab:public-expansion-ci}).

\begin{table}[ht]
\centering\small
\begin{tabular}{lllrrrr}
\toprule
Catalog & Encoder & Population & $\Delta$ exp. & Simultaneous 95\% & $\Delta$ RI & Simultaneous 95\% \\
\midrule
TOUCAN & Jina & Labeled & $+16.12$ & $[9.00, 23.56]$ & $+20.68$ & $[13.44, 27.55]$ \\
TOUCAN & Jina & Primary held-out & $-8.87$ & $[-14.89, -3.11]$ & $-6.95$ & $[-18.70, 4.80]$ \\
TOUCAN & Jina & All zero-support & $-7.68$ & $[-11.94, -3.66]$ & $-6.51$ & $[-13.05, -0.27]$ \\
\midrule
TOUCAN & Nomic & Labeled & $+8.50$ & $[3.65, 14.79]$ & $+7.19$ & $[1.86, 13.13]$ \\
TOUCAN & Nomic & Primary held-out & $-2.24$ & $[-6.67, 2.82]$ & $-1.48$ & $[-8.43, 7.51]$ \\
TOUCAN & Nomic & All zero-support & $-2.46$ & $[-5.46, 0.31]$ & $-3.95$ & $[-8.38, 0.73]$ \\
\midrule
APIBench & Jina & Labeled & $+1.02$ & $[-1.43, 3.68]$ & $+1.66$ & $[-1.66, 5.15]$ \\
APIBench & Jina & Held-out & $+1.11$ & $[-0.53, 2.83]$ & $-2.63$ & $[-5.52, 0.13]$ \\
\midrule
APIBench & Nomic & Labeled & $+4.12$ & $[0.55, 8.07]$ & $+1.75$ & $[-1.96, 5.82]$ \\
APIBench & Nomic & Held-out & $-0.64$ & $[-3.20, 1.92]$ & $-0.09$ & $[-2.26, 2.05]$ \\
\bottomrule
\end{tabular}

\caption{\method{} minus each public arm, Hit@5 points, with simultaneous 95\% intervals over the two encoders. exp.: public expansion; RI: Re-Invoke-style arm. Positive values favour \method{}.}
\label{tab:public-expansion-ci}
\end{table}

The combined index is post hoc: we designed it after seeing the public-expansion results in Tables~\ref{tab:fresh-test} and~\ref{tab:apibench}. In each release, a tool whose noisy count passes the threshold keeps its \method{} requests. Every other tool receives its requests from one public arm. All tools are scored with the per-field rule, with one $w$ tuned on development data and the threshold unchanged. It uses no new generation, training or privacy budget. On TOUCAN its indexes hold 10,346 to 22,204 requests, against about 7,300 for \method{}; on APIBench they hold 12,028 to 12,599, against about 9,500. Table~\ref{tab:combined-index} reports every population. With public-expansion requests, the combined index is below public expansion on every TOUCAN held-out population, by 4.87 to 12.04 points. On APIBench it is below \method{} on held-out queries in all four settings. With Re-Invoke-style requests, the combined index is below that arm in point estimate on all four TOUCAN held-out populations, and two of the four intervals exclude zero. One possible reason for the held-out shortfall is that tools with many \method{} requests get more chances to match a query under the maximum-similarity usage score than tools with 3 to 10 public requests. We did not test this.

\begin{table}[ht]
\centering\scriptsize
\setlength{\tabcolsep}{3.5pt}
\begin{tabular}{llllrrr}
\toprule
Catalog & Encoder & Public source & Population & Hit@5 & $-$ \method{} & $-$ public source \\
\midrule
TOUCAN & Jina & exp. & Labeled & 62.11 & $+4.76$ $[0.93, 8.28]$ & $+20.88$ $[15.02, 27.37]$ \\
TOUCAN & Jina & exp. & Primary held-out & 42.97 & $-1.30$ $[-8.13, 3.37]$ & $-10.18$ $[-16.45, -4.61]$ \\
TOUCAN & Jina & exp. & All zero-support & 51.09 & $-4.37$ $[-9.29, 0.38]$ & $-12.04$ $[-16.62, -7.69]$ \\
TOUCAN & Jina & RI & Labeled & 67.23 & $+9.89$ $[7.36, 12.52]$ & $+30.56$ $[23.57, 37.54]$ \\
TOUCAN & Jina & RI & Primary held-out & 44.98 & $+0.70$ $[-7.74, 6.92]$ & $-6.25$ $[-14.67, 2.50]$ \\
TOUCAN & Jina & RI & All zero-support & 53.93 & $-1.52$ $[-7.15, 3.89]$ & $-8.03$ $[-13.24, -2.90]$ \\
\midrule
TOUCAN & Nomic & exp. & Labeled & 68.40 & $+8.17$ $[5.47, 11.67]$ & $+16.66$ $[11.69, 24.82]$ \\
TOUCAN & Nomic & exp. & Primary held-out & 60.62 & $-2.64$ $[-10.21, 1.78]$ & $-4.87$ $[-9.03, -1.19]$ \\
TOUCAN & Nomic & exp. & All zero-support & 65.59 & $-3.47$ $[-6.88, -0.69]$ & $-5.92$ $[-9.41, -3.01]$ \\
TOUCAN & Nomic & RI & Labeled & 65.80 & $+5.57$ $[3.34, 8.49]$ & $+12.76$ $[7.45, 19.87]$ \\
TOUCAN & Nomic & RI & Primary held-out & 64.33 & $+1.07$ $[-5.68, 5.03]$ & $-0.40$ $[-4.78, 3.72]$ \\
TOUCAN & Nomic & RI & All zero-support & 69.50 & $+0.44$ $[-2.54, 2.71]$ & $-3.51$ $[-7.61, -0.23]$ \\
\midrule
APIBench & Jina & exp. & Labeled & 30.56 & $+0.86$ $[-0.33, 2.08]$ & $+1.88$ $[-0.31, 4.10]$ \\
APIBench & Jina & exp. & Held-out & 21.84 & $-4.99$ $[-6.65, -3.51]$ & $-3.88$ $[-5.31, -2.50]$ \\
APIBench & Jina & RI & Labeled & 33.03 & $+3.33$ $[2.18, 4.52]$ & $+4.99$ $[1.45, 8.75]$ \\
APIBench & Jina & RI & Held-out & 21.46 & $-5.38$ $[-7.26, -3.57]$ & $-8.01$ $[-10.78, -5.38]$ \\
\midrule
APIBench & Nomic & exp. & Labeled & 35.20 & $+3.69$ $[1.88, 5.69]$ & $+7.81$ $[3.75, 12.29]$ \\
APIBench & Nomic & exp. & Held-out & 24.29 & $-4.13$ $[-6.27, -2.16]$ & $-4.77$ $[-6.58, -3.08]$ \\
APIBench & Nomic & RI & Labeled & 34.09 & $+2.58$ $[0.79, 4.38]$ & $+4.33$ $[0.97, 7.85]$ \\
APIBench & Nomic & RI & Held-out & 26.03 & $-2.38$ $[-4.39, -0.55]$ & $-2.47$ $[-4.51, -0.62]$ \\
\bottomrule
\end{tabular}

\caption{The post hoc combined index raises labeled Hit@5 above \method{} but stays below its public source on held-out queries in point estimate. Columns marked $-$ give the combined index minus \method{} and minus its public source. Hit@5 (\%) and differences in points with simultaneous 95\% intervals over the two encoders. Public source: public expansion (exp.) or the Re-Invoke-style requests (RI). Designed after the public-expansion results were seen.}
\label{tab:combined-index}
\end{table}

\section{Count-weight sweep on TOUCAN}\label{app:count-sweep}
On Jina, \method{} is well above the curve. The count weight that matches its held-out Hit@5 ($\lambda=0.08$) reaches 39.15\% labeled against 57.35\% for \method{}, and no tested weight reaches its labeled score. On Nomic the margin is smaller. At equal labeled Hit@5 ($\lambda=0.14$, 60.33\%), counts reach 60.23\% held-out against 63.25\%.

Against the post hoc full-budget rule at $\lambda=0.3$, \method{} differs on labeled queries by $+3.26$ points on Jina, descriptive 95\% interval $[-1.67, 8.31]$, and $-6.48$ on Nomic, $[-10.16, -3.47]$, and leads on primary held-out queries by 9.85, $[6.52, 14.27]$, and 9.94, $[5.59, 15.70]$.

On Nomic we also chose two weights on development data. One matches \method{}'s development labeled score, the other its held-out score. At the first ($\lambda=0.17$), counts lose 4.25 held-out points against \method{}, interval $[-8.72, -0.87]$, and differ by $+1.67$ labeled points, $[-1.40, 5.33]$. At the second ($\lambda=0.02$), counts lose 8.12 labeled points, $[-10.98, -5.13]$, and differ by $+1.68$ held-out points, $[-1.74, 6.43]$.

Figure~\ref{fig:tradeoff} in Section~\ref{sec:counts} shows the post hoc sweep of the count weight.

\section{Held-out tools exposed to generated usage}\label{app:exposure}
The count noise gives some tools without records a positive count, and those tools receive generated usage. At $\varepsilon=4$ the selected thresholds give generated usage to 257 to 266 of the 1,602 tools without records on Jina ($\tau=1$) and to 686 to 703 on Nomic ($\tau=0$), across releases (Appendix~\ref{app:noise}, Table~\ref{tab:first-test-allocation}). All 81 tools with at least 20 records receive generated usage in every release. A descriptive split separates held-out queries whose gold tools received generated usage (exposed) from the rest. We look at the 658 held-out queries with a single gold tool, over 80 tools. At $\varepsilon=4$, exposed queries change by $-17.60$, $-7.61$ and $+7.08$ points on Jina across the three releases, and by $-12.17$, $-5.65$ and $-5.65$ points on Nomic. Exposure is set by noise, not assigned at random, so this association does not identify a causal effect. Exposed single-gold changes are negative in 20 of 24 private and 22 of 24 public-generator encoder and release conditions, which are not independent releases. On Jina, unexposed held-out tools also lose some retrieval. One possible path is that tools with generated usage move above them. We did not test this path.

\section{APIBench diagnostics}\label{app:apibench-diag}
In the APIBench pool, one sentence-encoder call has 1,936 records, while the median tool has 8.

\paragraph{Private and public generators on held-out queries.}
The protocol also compared \method{} with the public generator on held-out queries. \method{} is ahead by $+2.09$ points on Jina, simultaneous interval $[1.11, 3.26]$, and $+2.14$ on Nomic, $[1.14, 3.22]$. Most held-out gold tools receive no generated requests in either arm, so this difference reflects how the competing tools rank; we did not analyse it further.

\paragraph{Gain by hub.}
By hub, the gain is largest on TensorFlow Hub, where the copied entries give tools many records: $+6.90$ points on Jina and $+16.22$ on Nomic. The hubs also differ in other ways, so this does not isolate the number of records. HuggingFace tools have about the same number of records each, and there the gain is $+2.45$ points on Jina, descriptive 95\% interval $[0.53, 4.35]$, and $+1.93$ on Nomic, $[-0.37, 4.37]$. On TorchHub we detect no gain. These per-hub results are descriptive.

\paragraph{Count-only rule.}
In Table~\ref{tab:apibench}, development data selected count weight 0 in every case except Count (full) on Nomic (0.01). On development data, the count weight that matches \method{}'s held-out Hit@5 is zero for both encoders and both count budgets. At that weight the count rule is documentation, so this comparison repeats the labeled gain above. The informative comparison is the other one. At the weight that matches \method{}'s development labeled Hit@5 (0.16 to 0.29), \method{} is ahead on held-out queries by 9.40 to 13.42 points, and every simultaneous interval excludes zero. Across the full sweep of weights from 0 to 0.30, counts exceed \method{} on labeled queries only near $\lambda=0.3$, at 30.88\% on Jina and 32.28\% on Nomic with the full budget. At that weight, held-out Hit@5 falls to 12.51\% and 14.76\%. On this catalog the sweep was planned before scoring.

\paragraph{Where the held-out gain comes from.}
Held-out Hit@5 rises by 3.40 points on Jina and 3.06 on Nomic, and both intervals exclude zero. On TOUCAN the point estimates had the opposite sign, so after scoring we traced how this gain arises. Most held-out gold tools received no generated usage above the threshold. Their scores are identical under documentation and \method{}, and so are the scores of every competitor without generated usage. Over these 11,083 query and release pairs, the average number of tools with generated usage ranked above the gold tool falls from 68.73 to 63.68 on Jina and from 73.79 to 67.19 on Nomic. Hit@5 rises by 3.42 and 3.78 points. The 7,447 pairs whose gold tool had a rounded noisy count of zero show the same pattern. For these pairs, every rank change therefore comes from tools with generated usage whose scores changed. On TOUCAN we suggested, without testing it, that such tools may move above held-out tools (Appendix~\ref{app:exposure}). On APIBench they move from above the gold tool to below it more often than the reverse. This analysis does not separate the $(1-w)$ factor on their documentation term from the usage term, and we did not run it on TOUCAN. It does not show that \method{} represents held-out tools better. Where a held-out tool did receive generated usage through count noise, results differ by encoder: $+3.27$ points on Jina and $-0.83$ on Nomic, with Nomic losing in four of six releases.

\section{Generated-text overlap and released-output audits}
\label{app:release-audits}

These descriptive audits use the twelve first-test releases and their public-generator controls.

\paragraph{Text overlap.}
Among 111,397 indexed requests per generator arm, private training,
development and first-test near-copy counts are 37, 169 and 156; public
counts are 7, 1 and 12. A near-copy has cosine at least 0.95 under the Jina encoder, which we use for this audit with both retrieval encoders. Reference sets can
overlap and counts are not additive. The private training fraction ranges
from zero to 0.145\% per release. Byte-exact and normalized exact matches to
each reference set are zero for both generators. No output was filtered.
These checks do not rule out memorization or measure extraction risk.

\paragraph{Record membership.}
Two attacks score a candidate by its nearest synthetic-request cosine under
the Jina encoder: once over all requests, once over requests assigned to its
gold tools. Larger similarity always predicts membership; the direction is
not flipped after observing results. We form 3,418 unique member/nonmember
pairs matched on the entire retained gold-tool set, with 2,803 evaluation
pairs across 269 tool clusters (Table~\ref{tab:membership-audit}). A separate cluster split fits
thresholds. Multi-tool queries can share other gold labels across groups.
Unmatched records are excluded, so results do not cover every pool record.

\begin{table}[ht]
\centering\small
\begin{tabular}{lrlr}
\toprule
Generator & $\varepsilon$ & Global NN [95\% CI] & Gold-matched NN [95\% CI] \\
\midrule
Private & 1 & 0.493 [0.483, 0.504] & 0.496 [0.486, 0.505] \\
Private & 2 & 0.495 [0.484, 0.506] & 0.493 [0.484, 0.503] \\
Private & 4 & 0.492 [0.481, 0.503] & 0.493 [0.484, 0.502] \\
Private & 8 & 0.492 [0.481, 0.503] & 0.491 [0.483, 0.502] \\
Public & 1 & 0.494 [0.483, 0.506] & 0.495 [0.487, 0.504] \\
Public & 2 & 0.495 [0.484, 0.506] & 0.497 [0.488, 0.506] \\
Public & 4 & 0.497 [0.487, 0.508] & 0.495 [0.487, 0.504] \\
Public & 8 & 0.495 [0.485, 0.505] & 0.496 [0.488, 0.505] \\
\bottomrule
\end{tabular}

\caption{Released-output membership AUROC and descriptive 95\% intervals,
resampling three whole releases and matched-pair tool clusters. Global NN: nearest-neighbour cosine over all synthetic requests. Gold-matched NN: only over requests of the candidate's gold tools. AUROC: area under the ROC curve.
All intervals include 0.5. Public controls reuse the private count release.}
\label{tab:membership-audit}
\end{table}

A positive control that matches against the raw records has AUROC 1.0. Filtering leaked queries from the usage pool affects how easy that control is. These two attacks show
little membership discrimination on the matched evaluation population.
They are not likelihood, white-box or extraction attacks, and do not
establish membership resistance or an empirical privacy budget. The
benchmark uses reproducible public noise seeds, not a production mechanism
with secret randomness. We make no user-level claim.

\paragraph{Cached-vector cost.}
On the shared server of Appendix~\ref{app:data}, using four NumPy threads, median scoring time over
five repeats on 256 queries is 0.145 ms/query for documentation
and 0.222 for \method{}'s per-field scoring. This includes dense cosine,
field aggregation and stable top-five ranking; it excludes query embedding,
training, generation, disk I/O, network and GPU utilization. It is neither
end-to-end latency nor an isolated-load or dollar-cost benchmark.

\paragraph{Count-only weight.}
The count-only score is $z^{\mathrm{doc}}_t(q)+\lambda\log(1+\max(\tilde c_t,0))$. We chose $\lambda$ from $\{0,0.01,0.03,0.1,0.3,1\}$ with the balanced development objective on the $\varepsilon=4$ calibration release, breaking ties toward smaller weights; it selects 0.3 for both encoders. Its development labeled and held-out Hit@5 are 60.06\% and 52.36\% on Jina and 71.00\% and 63.71\% on Nomic. It reuses the six second-test count releases without new noise and spends only the count share of the budget.

\end{document}